\documentclass[letterpaper,journal]{IEEEtran}
\usepackage{amsmath,amsfonts}
\usepackage{algorithmic}
\usepackage{algorithm}
\usepackage{array}
\usepackage[caption=false,font=normalsize,labelfont=sf,textfont=sf]{subfig}
\usepackage{textcomp}
\usepackage{stfloats}
\usepackage{url}
\usepackage{verbatim}
\usepackage{graphicx}
\usepackage{cite}
\usepackage{booktabs}
 \usepackage{multirow}
\usepackage{xcolor}
\usepackage{bbm}
\usepackage{tabularx}
\usepackage[table]{xcolor}
\usepackage{algorithm,algorithmic}


\definecolor{pure_purple}{RGB}{128, 0, 128}

\usepackage{pgfplots}
\pgfplotsset{compat=newest}

\begin{document}

\title{Unbiased Semantic Decoding with Vision Foundation Models for Few-shot Segmentation}

\author{Jin Wang, Bingfeng Zhang,~\IEEEmembership{Member,~IEEE}, Jian Pang, Weifeng Liu,~\IEEEmembership{Senior Member,~IEEE}, Baodi Liu, and Honglong Chen,~\IEEEmembership{Senior Member,~IEEE},
\thanks{This work was supported in part by the National Natural Science Foundation of China (Grant No. 62372468, 62301613), in part by the Shandong Natural Science Foundation (Grant No. ZR2023QF046, ZR2023MF008), in part by the Major Basic Research Projects in Shandong Province (Grant No.ZR2023ZD32), in part by the Taishan Scholar Program of Shandong (No. tsqn202306130).(\emph{Corresponding authors: Bingfeng Zhang; Weifeng Liu}.)\\
Jin Wang, Bingfeng Zhang, Jian Pang, Weifeng Liu, Baodi Liu and Honglong Chen are with the School of Control Science and Engineering, China University of Petroleum (East China), Qingdao, Shandong 266580, China. (e-mail: wangjin@s.upc.edu.cn; bingfeng.zhang@upc.edu.cn, bingfeng.z@outlook.com; jianpang@s.upc.edu.cn; liuwf@upc.edu.cn; liubaodi@upc.edu.cn; chenhl@upc.edu.cn).
}}


\markboth{IEEE TRANSACTIONS ON NEURAL NETWORKS AND LEARNING SYSTEMS}%
{Shell \MakeLowercase{\textit{et al.}}: A Sample Article Using IEEEtran.cls for IEEE Journals}


\maketitle
\begin{abstract}
Few-shot segmentation has garnered significant attention. Many recent approaches attempt to introduce the Segment Anything Model (SAM) to handle this task. With the strong generalization ability and rich object-specific extraction ability of the SAM model, such a solution shows great potential in few-shot segmentation.
However, the decoding process of SAM highly relies on accurate and explicit prompts, making previous approaches mainly focus on extracting prompts from the support set, which is insufficient to activate the generalization ability of SAM, and this design is easy to result in a biased decoding process when adapting to the unknown classes.
In this work, we propose an Unbiased Semantic Decoding (USD) strategy integrated with SAM, which extracts target information from both the support and query set simultaneously to perform consistent predictions guided by the semantics of the Contrastive Language-Image Pre-training (CLIP) model. Specifically, to enhance the unbiased semantic discrimination of SAM, we design two feature enhancement strategies that leverage the semantic alignment capability of CLIP to enrich the original SAM features, mainly including a global supplement at the image level to provide a generalize category indicate with support image and a local guidance at the pixel level to provide a useful target location with query image. Besides, to generate target-focused prompt embeddings, a learnable visual-text target prompt generator is proposed by interacting target text embeddings and clip visual features. Without requiring re-training of the vision foundation models, the features with semantic discrimination draw attention to the target region through the guidance of prompt with rich target information. Experiments on both the PASCAL-5$^{i}$ and COCO-20$^{i}$ show that our proposed method outperforms existing approaches by a clear margin and achieves new state-of-the-art performances. The code is available on https://github.com/vangjin/USD.
\end{abstract}


\section{Introduction}
\IEEEPARstart{S}{emantic} segmentation has advanced significantly with the emergence of deep learning techniques~\cite{jiang2024prototypical, yang2021part, guo2018review, lang2024toward, yuan2025distance}. However, training robust semantic segmentation models requires large-scale pixel-level annotated datasets, which are labor-intensive~\cite{pan2024cc4s} and costly to produce~\cite{gao2022large}. The exponential growth of data and the inherent limitations of manual annotation further exacerbate the challenge, rendering traditional semantic segmentation models inadequate for open-world scenarios and hindering their broader application. To address these issues, few-shot segmentation (FSS) has been introduced~\cite{he2022learning, luo2024layer}. This paradigm enhances model generalization by transferring knowledge from closed-world training to open-world inference using minimal annotation information, thereby reducing reliance on annotated data. In FSS, data are partitioned into support set and query set: the model leverages few annotations from the support set to segment objects of novel classes in query images, enabling efficient adaptation to unseen tasks.

\begin{figure*}[!t]
	\centering
	\includegraphics[width=\textwidth]{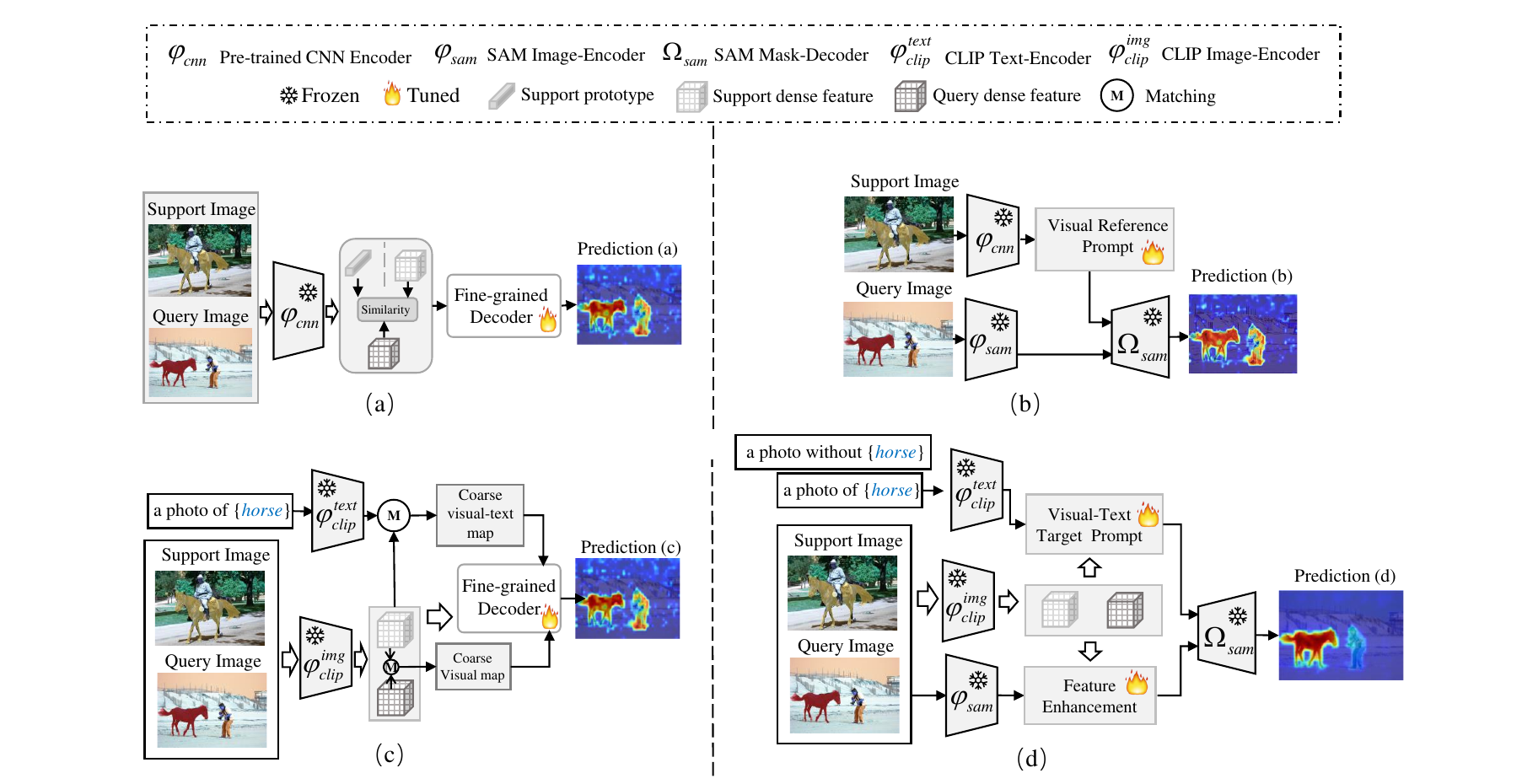}
	\caption{Comparison of different FSS methods. (a) Existing methods of prototype-level or pixel-level matching FSS methods, obtain predictions by designing fine-grained decoders. (b) The framework of existing SAM-based FSS methods, which primarily aims at extracting visual features from support image using ImageNet pre-trained CNN weights for prompt generation. (c) The framework of existing CLIP-based FSS methods, which primarily obtain coarse map generation by visual-visual matching or visual-text matching. (d) Our proposed strategy integrates visual-text prompt generation from both support and query images with a frozen decoder, and enhances the semantic information of SAM features by complementing them with CLIP features.}
	\label{fig:1}
\end{figure*}

Existing few-shot segmentation (FSS) methods are broadly categorized into two paradigms: pixel-level dense matching~\cite{hong2022cost, shi2022dense, bi2024agmtr, xu2024eliminating, zhang2023rpmg, tian2020prior} and prototype-level matching~\cite{jiang2024prototypical, he2024apseg, shaban2017one, zhang2021self, lang2022learning, lang2023base, fan2022self, lang2024few}, as shown in Fig.~\ref{fig:1} (a). Pixel-level dense matching methods establish correlations by comparing individual pixels between query and support images, prototype-level matching methods aggregate support features into a prototypical representation and perform similarity matching with query dense features. However, whether it is a pixel-level approach or a prototype-level approach, they both require finely designed decoders to capture the correlation between the support samples and the query samples. Such a fine-grained decoder usually needs a large number of learnable parameters, leading to bias of the model to the base class during training, such as the `person' class in the prediction (a). This leads to difficulties in adapting the model to the characteristics of the new class, which further raises the issue of insufficient generalization to the new class, producing an inaccurate segmentation of the new class, as in the case of the `horse' class in prediction (a).

Recently, some methods have integrated vision foundation models~\cite{lin2023clip, sun2024vrp, li2025clipsam} into FSS to enhance generalization under both paradigms. Among these, SAM~\cite{kirillov2023segment} is the most widely adopted. SAM is pre-trained on the SA-1B dataset, which contains over 11 million images and 1 billion segmentation masks. Additionally, its interactive prompt design and feature consistent representation enable precise segmentation. By leveraging the unique interactive prompts of SAM, such as points, boxes, and masks, FSS models can obtain more informative target-specific prompts, improving object localization and enhancing segmentation accuracy.

SAM-based few-shot segmentation methods have achieved significant progress, as shown in Fig.~\ref{fig:1} (b), they first extract support information from another CNN pre-training models~\cite{deng2009imagenet} as visual prompts, then the visual prompts are used directly to segment the query features extracted from SAM. However, in few-shot tasks where the support samples are limited and the query samples are large and diverse, the same target class in the support samples and query samples may have different positions, colors, shapes, and so on, resulting in intra-class differences. This leads to existing SAM-based methods suffering from some problems: 1) \textbf{limited semantic feature representation}: decoded features only from SAM leads to missing semantic guidance, since SAM is designed to focus on achieving arbitrary targets based on user-provided interactive prompts~\cite{kirillov2023segment}, which causes the models to excessively prioritize generalization ability and prompt responsiveness, leading to an insufficient in-depth understanding of semantic information; 2) \textbf{inaccurate semantic discrimination prompts}: the support prompts generated from pre-trained CNN weights cannot effectively activate the SAM query features, the main reason for this is the poor semantic consistency of pre-trained CNN features of the same class and the significant difference in semantic space between pre-trained CNN features and SAM features.

Based on these two reasons, SAM-based methods are difficult to adapt to new classes and suffer from base classes bias. To address the above problems, we propose an Unbiased Semantic Decoding (USD) method with vision foundation models for few-shot segmentation, aiming at further activating the generalization of the original SAM model with the semantic help of CLIP. The forced alignment strategy of the text-image pairs makes the CLIP model sensitive to semantic class concepts, enabling it to capture more semantic useful information from target classes~\cite{lin2023clip}. 
As shown in Fig.~\ref{fig:1} (c), we first propose a feature enhancement strategy at both image-level and pixel-level to improve the effective semantic content and consistency of the decoded features, abundant feature representation improves the upper bound of SAM segmentation. Secondly, considering SAM is sensitive to prompts and for effective activation of enriched features, a multi-modal target prompt is designed to obtain useful semantic representations. Specifically, we propose two kinds of feature enhancement strategies, the first one is Global Supplement Module (GSM), through a light learnable Multilayer Perceptron, the CLIP visual support feature and query feature are mapped in image-level to the feature space of SAM, enabling the model with the ability to recognize the semantic category of the whole image. Since GSM provides semantic complements at the image-level, it lacks spatial localization capabilities and local detail information, which results in insufficient guidance for intensive segmentation prediction tasks. The other feature enhancement strategy Local Guidance Module (LGM) is proposed to focus on the pixel-level probability guidance, aiming to provide the localization and detailed information of the target area based on the semantic affinity of clip visual features. After the application of image-level and pixel-level feature enhancement strategies, the consistency and semantic representation of features are significantly improved.

Furthermore, under the framework of SAM, the decoder needs prompts to understand the image content and the accuracy of the prompt is vital for the correct segmentation results~\cite{lyu2024superpixel, he2024apseg, zhangpersonalize}. To enhance the valid target semantic information of the prompt, we propose Visual-Text target Prompt Generator (VTPG) to force an alignment between the visual information from both support image and query image with the text information of the target class. Through mining the available target information from the query image itself, the intra-class difference between support and query is alleviated, and a query-aware prompt is obtained to further activate the generalization of SAM. Without retraining SAM and CLIP, our proposed USD method can effectively overcome class bias by adding only a small number of learnable parameters for the model, effectively improving the performance.

Our Contributions are summarized as follows:

1) We observe that the previous SAM-based FSS methods suffer from bias towards the base class, and we propose a new joint optimization strategy combining CLIP and SAM to build an unbiased semantic decoding process.

2) Without retraining the vision foundation models, we propose image-level and pixel-level feature enhancement techniques and develop a multi-modal prompts mechanism to achieve the effective activation of target features.

3) Comprehensive experiments show that our proposed method has a significant improvement over existing FSS methods on both PASCAL-5$^{i}$~\cite{shaban2017one} and COCO-20$^{i}$~\cite{nguyen2019feature} datasets and achieves state-of-the-art performance.

\section{Related Work}
\subsection{Few-Shot Segmentation}
Few-shot Segmentation aims at performing a dense mask prediction for unknown query image with a few annotation support samples. Most few-shot segmentation methods follow the meta-learning paradigm. According the match mode of support samples and query samples, existing few-shot segmentation methods can be divided into two types: prototype-level matching~\cite{peng2023hierarchical, liu2022dynamic, lang2024few} and pixel-level matching~\cite{yang2020brinet, hong2022cost, shi2022dense}. Prototype-level matching methods aim to extract target information from the supporting images as prototype vectors for intensive matching or similarity computation with the query image. The pioneering work OSL~\cite{shaban2017one} formally introduces the few-shot segmentation (FSS) task, employing parallel networks to model the relationship between support images and the query image. Building upon this foundation, prototype-level matching work SG-One~\cite{zhang2020sg} refines the parallel architecture and adopts a masked average pooling strategy to obtain more robust prototype-level representations. Recognizing that such prototypes inevitably suffer from information loss, SCL~\cite{zhang2021self} addresses this limitation by mining the missing details from the discrepancy between two predictions of the support image. To further enhance foreground recognition under prototype guidance, BAM~\cite{lang2022learning,lang2023base} introduces a base learner to suppress erroneous novel-class predictions from the meta-learner. Complementarily, FBINet~\cite{huang2025fbinet} proposes to jointly integrate foreground and background semantic knowledge in order to enhance the foreground location and suppress background interference. SSP~\cite{fan2022self} extends prototype utilization by performing a second prediction between the support prototype and the query image, enabling the mining of more informative query features. Recent approaches further exploit external priors under prototype guidance: LLaFS~\cite{zhu2024llafs} leverages the knowledge of large language models to enrich prototype representations, while MGCL~\cite{li2024mask} incorporates SAM to maintain feature consistency between foreground and background masks. In parallel, pixel-level matching methods focus on dense correspondences between the support and query images. PFENet~\cite{tian2020prior} generates target priors by pixel-to-pixel interaction of high-level visual features, while HDMNet~\cite{peng2023hierarchical} captures richer contextual dependencies by mining pixel-level correlations via a transformer-based architecture. Pushing this interaction further, DiffewS~\cite{zhu2024unleashing} employs a diffusion model to establish deep, iterative interactions between the support and query images, providing stronger supervision for pixel-level matching. Whether prototype-level matches or pixel-level matches, they require heavy decoder for decoding, which tends to cause overfitting and exacerbates the problem of bias towards base classes. We propose an unbiased semantic decoding with a frozen decoder to mitigate this bias.

\subsection{SAM-based FSS}
The Segment Anything Model (SAM)~\cite{kirillov2023segment} comprises three core components: an image encoder, a prompt encoder, and a mask decoder. Owing to its strong generalization capability, SAM has shown remarkable potential not only in few-shot learning tasks~\cite{zhang2023personalize, bai2024fs, li2024mask}, but also in zero-shot learning scenarios~\cite{li2025clipsam, roy2023sam, deng2023segment, gao2025combining}. Recently, the foreground-background segmentation paradigm adopted by SAM has been recognized as highly compatible with the foreground--background recognition objective in FSS, leading to its widespread adoption in this domain. Pioneering this integration, VRP-SAM~\cite{sun2024vrp} first introduces SAM into FSS by using annotated support images as prompts to segment specific objects in the query image. To generate more informative prompts, FCP~\cite{park2025foreground} constructs reliable prompts based on the relationship between support and query prototypes, while APSeg~\cite{he2024apseg} further automates this process by designing an automatic prompt generation mechanism for SAM. Departing from prompt enhancement, CAT-SAM~\cite{xiao2024cat} proposes a conditional tuning network with lightweight learnable parameters for efficient few-shot adaptation of SAM. Although SAM-based FSS methods have made notable progress, most focus on prompt design or decoder fine-tuning, overlooking two key issues: (1) SAM features contain limited semantic information, and (2) intra-class variations remain unresolved. We address them by enriching SAM features with semantically rich CLIP embeddings for stronger feature consistency and aligning both support and query features with text embeddings to reduce intra-class variation.

\subsection{CLIP-based FSS}
Contrastive Language-Image Pretraining (CLIP)~\cite{radford2021learning} is a pioneering framework that aligns text features from a text encoder with visual features from a visual encoder via contrastive learning. This alignment enables strong generalization across diverse downstream tasks, such as object detection~\cite{ju2022adaptive, jiang2024t} and semantic segmentation~\cite{lin2023clip, yang2024multi, chen2024visual}. Building on CLIP, PGMA-Net~\cite{chen2024visual} introduces class-agnostic priors, obtained by matching text embeddings with visual features. PAT~\cite{bi2024prompt} further proposes a dynamic text class-aware prompting to guide the encoder segment target objectives. While these methods leverage text features, they primarily use foreground representations as coarse semantic guidance. In contrast, we incorporate text information into the decoding stage to diffuse semantics throughout the features, and explicitly model both foreground and background representations to better suppress base-class interference.

\section{Task Description}
 Few-shot segmentation aims at transferring the segmentation ability learned from the known base classes to the unknown novel class. Most existing few-shot segmentation approaches follow the framework of meta-learning. In the training stage, the model undergoes optimization via multiple meta-learning tasks, and the efficacy is evaluated in the testing stage. Given a dataset $D$, dividing it into a training set ${D_{train}}$ and a test set ${D_{test}}$, the class set in the training set ${C_{train}}$ and the class set in the test set ${C_{test}}$ are completely distinct, with no overlapping classes. The model is designed to transfer the knowledge in ${D_{train}}$ with limited labeled data to the diverse ${D_{test}}$. Both training set ${D_{train}}$ and test set ${D_{test}}$ containing support set $S$ and query set $Q$, support set $S$ contains $K$ samples $S=\{S_1,S_2,\,\ldots,S_K\}$, each $S_i$ contains an image-mask pair $\{I_s, M_s\}$ and query set $Q$ contains $N$ samples $Q=\{Q_1,Q_2,\,\ldots,Q_N\}$, each $Q_i$ contains an image-mask pair $\{I_q, M_q\}$. During training, the few-shot model performs prediction for query image $I_q$ with the guidance of the support set $S$ and the model is optimized under the supervision of the query mask $M_q$. During inference, the model is fixed and the performance will be acquired by a direct prediction on the test set ${D_{test}}$ in the same way as training. In this phase, query query mask $M_q$ is only utilized to evaluate the performance of the model.
\section{Method}
\subsection{Method Overview}
To alleviate the class bias problem caused by limited support samples and diverse query samples in FSS, we propose to perform an unbiased semantic decoding process on SAM with the semantic guidance from CLIP, which mainly includes enhancing the semantic distinguishing ability of features and improving the utilization of target information. Firstly, we augment the original SAM features with CLIP-derived semantic-aligned features at both image and pixel levels to strengthen semantic distinctions of features. Furthermore, for the SAM decoder, an accurate prompt is necessary~\cite{deng2023segment}, so we interact with the target text embedding and the target visual features to obtain a more informative multi-modal target prompt. Fig.~\ref{Fig2} shows our framework of the one-shot case with the following steps:
\begin{figure*}[!t]
\centering
\includegraphics[width=1.0\textwidth]{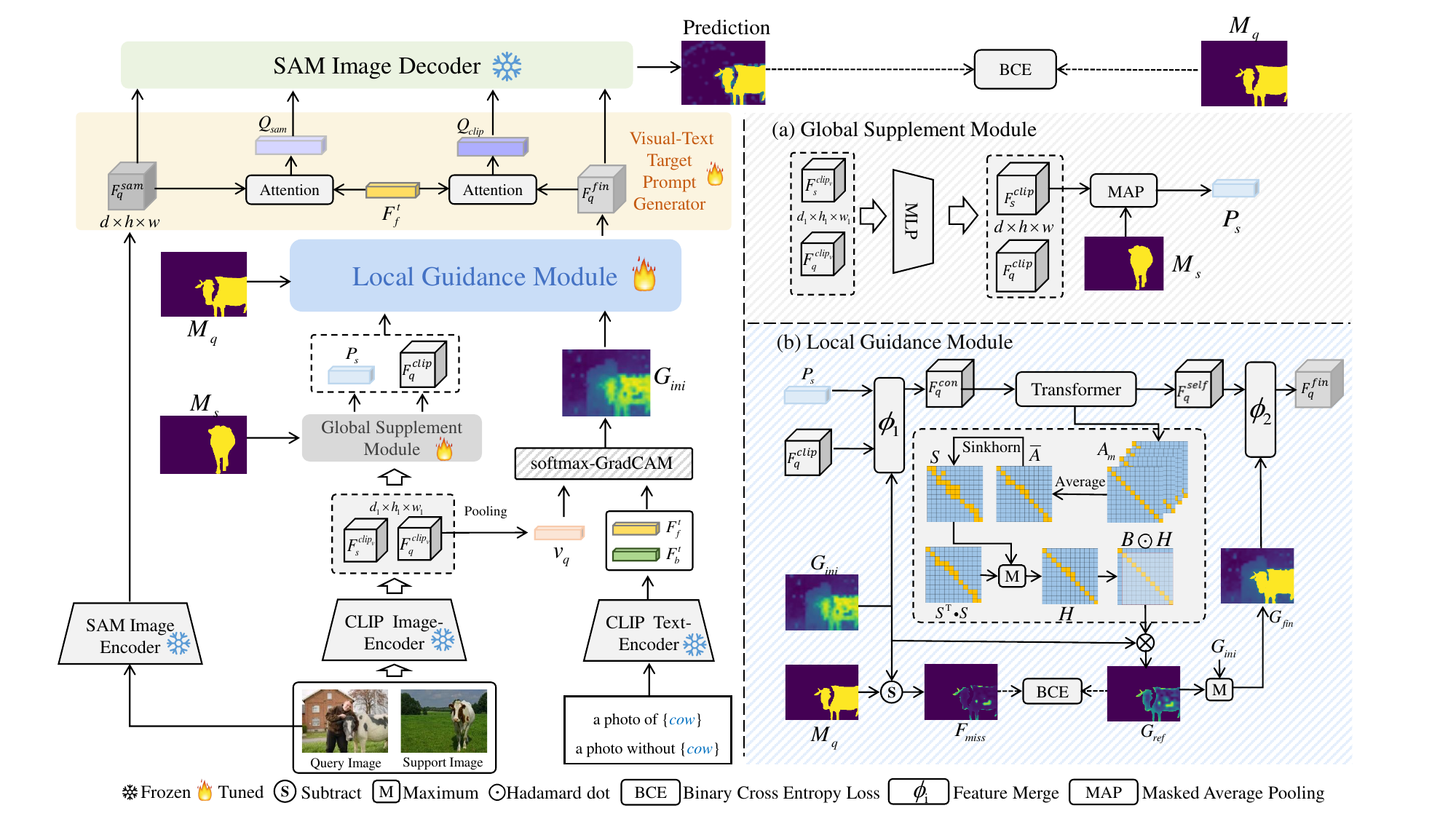}
\caption{Overview of the proposed Unbiased Semantic Decoding method under 1-shot setting. We design two feature enhancement strategies, image-level GSM and pixel-level LGM, to enhance the semantic information of SAM features using semantically rich CLIP features. The GSM enhances SAM features with image-level semantic information by mapping CLIP visual features to SAM space. The LGM enriches SAM features by mining semantic correlations between pixels to obtain pixel-level category representations. Finally, without any re-training of the SAM and CLIP models, VTPG is proposed to generate a multi-modal target to further activate the target regions.} 
\label{Fig2}
\end{figure*}

\begin{enumerate}
\item Given a collection of support images, query images, and their associated text prompts, the query and support images are processed through the CLIP image encoder to produce CLIP visual features for both support and query images. In addition, the query image is encoded by the SAM image encoder to generate SAM query features. Concurrently, the target and non-target text prompts are fed into the CLIP text encoder to generate two distinct text embeddings.

\item Then, the extracted CLIP visual support features and query features are aligned to the SAM representation space, through a lightweight network integrated within the proposed Global Supplementary Module (GSM), facilitating the acquisition of image-level aligned features.

\item Meanwhile, two text embeddings, the CLIP visual query features from the frozen CLIP image encoder and the output of GSM are input to the local guidance module (LGM). Through this, the final clip features and multi-modal pixel-level probability maps are generated based on the pixel-pixel relationships.

\item After that, the final clip features, the SAM query features, and the target text embeddings are sent to the visual-text target prompt generator (VTPG) to generate a multi-modal prompt. 
\end{enumerate}

Finally, the final CLIP feature is used to compensate the original SAM features, and the compensated features are sent to the SAM decoder to generate the final prediction with the multi-modal prompts.
\subsection{Global Supplement Module}
Existing few-shot segmentation methods based on SAM models primarily extract informative prompts from support images. However, the limited number of support images and the intra-class differences between support and query images constrain the effectiveness of these prompts. Moreover, the foreground-background training strategy~\cite{kirillov2023segment} causes SAM to generate numerous masks lacking semantic significance. To produce semantically meaningful masks, we enhance the semantic discriminative capability of the features. Specifically, we design an image-level global supplement module to map semantically informative CLIP features into the SAM space.

Firstly, the support image $I_s$ and the query image $I_q$ are sent to the CLIP image encoder, the corresponding CLIP visual query feature $F_{q}^{clip_v} \in \mathbb{R}^{d_{1} \times h_{1} \times w_{1}}$ and CLIP visual support feature $F_{s}^{clip_v} \in \mathbb{R}^{d_{1} \times h_{1} \times w_{1}}$ are generated after removing the class token. Meanwhile, the query image $I_q$ is sent to the SAM image encoder to obtain the SAM query feature $F_q^{sam} \in \mathbb{R}^{d \times h \times w}$. Subsequently, the CLIP visual features are interpolated to match the size $h \times w$ of $F_q^{sam}$ and mapped into the feature space of SAM by learnable MLP:
\begin{equation}
\begin{aligned}
F_{s}^{clip}= MLP_1(F_{s}^{clip_v}),
\end{aligned}
\end{equation}
\begin{equation}
\begin{aligned}
F_{q}^{clip}= MLP_2(F_{q}^{clip_v}),
\end{aligned}
\end{equation}
where the two learnable MLPs are composed of a convolutional layer followed by a RELU activation function and they are parametrically independent.

After aligning the clip visual features to the SAM space, for the consistent fusion later, the support prototype is generated with the binary support mask $M_s$ using the masked average-pooling~\cite{zhang2020sg, zhang2019canet, zhu2021self}:
\begin{equation}\label{eq0}
\begin{aligned}
P_s= \frac{\sum_{i=1}^{hw} F_{s}^{clip}(i) \cdot \mathbbm{1} (M_i^{s}=1)}{\sum_{i=1}^{hw} \mathbbm{1} (M_i^{s}=1)}
\end{aligned},
\end{equation}
where $i$ represents the pixel index, $h$ and $w$ are the height and width of the feature map, respectively. $\mathbbm{1}$ is the indicator function, $M_i^{s}=1$ indicates the $i$-th pixel belongs to class $c$. Note that $M^{s}$ is reshaped as the same size with $F_s^{clip}$. 

\subsection{Local Guidance Module}
The proposed GSM above maps the semantically informative CLIP features into the feature representation space of SAM, which enriches the semantic discriminative ability of features at the image level. However, this image-level semantic supplement is difficult to support intensive semantic segmentation tasks. In order to perform pixel-level semantic discrimination, we propose LGM to force the model to adaptively learn the semantic category probability of each pixel, which leads to a more useful target location.

The LGM consists of two parts: fixed local guidance generation and dynamic local guidance fine-tuning. The fixed local guidance is used to produce a pixel-level category indicating based on the visual-text features from the frozen CLIP. Moreover, dynamic local guidance fine-tuning is proposed to expand the localization area by learning pixel-level visual correlations dynamically. We design these two parts based on the following considerations: the forced alignment of frozen visual features and frozen text embedding restricts the extent of its region of interest, and the fixed local guidance is unable to dynamically expand its localization area, resulting in a limited effective localization area. By utilizing the fixed local guidance generation, the target region achieves an initial localization, after the fine-tuning, the effective target location area is dynamically expanded. Rich target information effectively mitigates model bias to base classes.

Suppose $F^t_f$ and $F^t_b$ represent the target (foreground) and non-target (background) text embeddings generated from the frozen CLIP text encoder with the description of ``a photo of \{\emph{target class}\}" and ``a photo without \{\emph{target class}\}". In order to obtain the effective pixel-level prior guidance, we first measure the similarity scores between the CLIP visual query features and the two CLIP text embeddings on a pixel-by-pixel basis, and obtain fixed local guidance based on softmax-GradCAM~\cite{lin2023clip} through the similarity score by: 
\begin{equation}
\begin{aligned}
S = softmax(\frac{v_q^\mathrm{T}F^t_i}{\|v_q\|\|F^t_i\|}/\tau), i \in \{ {f, b} \}
\end{aligned},
\end{equation}
where $\tau$ is a temperature parameter and $v_q$ is the pooling of the CLIP query visual features $F_{q}^{clip_{v}}$. By calculating the gradient of the similarity score $S$ over $F_{q}^{clip_{v}}$, we can get the weight of the foreground part, and the initial fixed local guidance $G_{ini}$ is obtained after weighting the $F_{q}^{clip_{v}}$. By suppressing the expression of background, $G_{ini}$ is able to focus more on effective target regions.

Then, we perform the consistent fusion through features and local guidance $G_{ini}$ to enhance the target consistency of the query feature for an efficient decoding process. The CLIP target consistent query feature is obtained by:
\begin{equation}
\begin{aligned}
F_q^{con}= \phi_1(P_s \oplus F_q^{clip} \oplus G_{ini})
\end{aligned},
\end{equation}
where $\phi_1$ means a light dimensionality reduction module, $\oplus$ means the cascade along the channel dimension and $P_s$ comes from Eq.~\ref{eq0}.

After that, we further propose to dynamically refine the initial fixed local guidance $G_{ini}$ by utilizing the pixel-level relationships within the CLIP consistent query feature $F_q^{con}$. Specifically, we propose to utilize a learnable self-attention mechanism to capture the global feature representation and the semantic affinity relationship of $F_q^{con}$ by:
\begin{equation}
\begin{aligned}
F_q^{self}, A_m= SelfAttn(F_q^{con})
\end{aligned},
\end{equation}
where $SelfAttn$ represents the self-attention mechanism~\cite{vaswani2017attention}, $F_q^{self}$ means the global features and $A_m$ means the attention maps from the $m$-th block. To acquire more useful semantic relationships for each image, we first compute the average attention map by:
\begin{equation}
\begin{aligned}
\overline{A} = \frac{1}{l} \sum_{m=n-l}^{n} A_m
\end{aligned},
\end{equation}
where $l$ and $n$ are the block number of the transformer and $l<n$. By summing and averaging the attention maps of the last $l$ layers, we end up with $\overline{A}$. To minimize the impact of background regions and maintain the essential structural details,  the average attention map $\overline{A}$ is aligned from rows and columns by Sinkhorn normalization~\cite{sinkhorn1964relationship} and get $S$, then, a high-order refinement matrix $H \in \mathbb{R}^{1 \times h \times w}$ is designed follows:
\begin{equation}
\begin{aligned}
H = max(S,(S\cdot S^{T}))
\end{aligned},
\end{equation}
where $max$ represents the maximum value at the corresponding pixel position.
We then utilize the refinement matrix $H$ to refine the initial fixed local guidance $G_{ini}$ by:
\begin{equation}
\begin{aligned}
G_{ref} =  B \odot H \cdot G_{ini}
\end{aligned},
\end{equation}
where $B$ is a box mask generated from $ G_{ini}$ following CLIP-ES~\cite{lin2023clip}, $\odot$ represents the Hadamard product.

After that, to focus on more target regions, the refined local guidance $G_{ref}$ learns the areas of the target foreground that $G_{ini}$ does not pay attention to. Firstly, the target foreground area ignored by $G_{ini}$ is extracted as a pseudo label by:
\begin{equation}
\begin{aligned}
F_{miss} = 
\begin{cases}
0, & \text{if } \text{$M_q$} = 0, \\
\text{$M_q$} - \text{$G_{ini}$}, & \text{otherwise}.
\end{cases}
\end{aligned}
\end{equation}
Then the refined local guidance is supervised by the missed target foreground $F_{miss}$ by:
\begin{equation}\label{eq1}
\begin{aligned}
loss_{ref} = BCE(G_{ref}, F_{miss})
\end{aligned},
\end{equation}
where $BCE$ means the binary cross entropy loss function, and the final guidance $G_{fin}$ is selected from the maximum value of the corresponding position from $G_{ref}$ and $G_{ini}$:
\begin{equation}
\begin{aligned}
G_{fin} = max(G_{ini}, G_{ref})
\end{aligned},
\end{equation}
then the second feature enhancement is performed to obtain the final CLIP query features sent to the decoder with richer target information:
\begin{equation}
\begin{aligned}
F_q^{fin} = \phi_2(F_q^{self} \oplus G_{fin}),
\end{aligned}
\end{equation}
where $\oplus$ means the cascade operation along the channel dimension, and $\phi_2$ is another dimensionality reduction network with the same structure as $\phi_1$, note that the local guidance $G_{fin}$ is resized as the same size as $F_q^{self}$.

\subsection{Visual-Text Target Prompt Generator}
The previously discussed global supplement module and local guidance module enhance the semantic representation of SAM features at the image and pixel levels, respectively. The SAM structure demands high-quality prompts during the decoding process. For FSS task, it is essential to provide unbiased and target-focused prompts since the tendency to overfit the base classes during training.

We propose the VTPG to align the useful target prior both from text-modal and visual-modal. After generating visual query features with stronger feature consistency through feature fusion, the VTPG combines textual context with visual features to capture semantic relationships and contextual information within images.

With the target text embeddings $F_f^{t}$ obtained by frozen CLIP text encoder, the CLIP target visual-text prompt $Q_{clip}$ can be obtained by:
\begin{equation}
\begin{aligned}
Q_{clip},\_ = SelfAttn(CrossAttn(fc(F_f^{t})), {F_q^{fin}})),
\end{aligned}
\end{equation}
where $SelfAttn$ means the self-attention, the $CrossAttn$ means the cross-attention and the $fc$ represents a fully connected layer used to align the channel dimension of text embeddings with the channel dimension of visual features. Meanwhile, in order to maintain the original segmentation capabilities of the SAM model, the SAM visual features are also used to interact with the CLIP text features to obtain the SAM visual-textual prompt:
\begin{equation}
\begin{aligned}
Q_{sam},\_ = SelfAttn(CrossAttn(fc(F_f^{t})), F_q^{sam})).
\end{aligned}
\end{equation}

\subsection{Training Loss}
Ultimately, without requiring retraining or fine-tuning of the SAM and CLIP models, semantically expressive features can be effectively activated through target-enriched prompts to produce unbiased semantic results both from CLIP features and SAM features, and the frozen SAM decoder $D$ is utilized:
\begin{equation}
\begin{aligned}
P_{clip}, P_{sam} = D(F_q^{fin}, F_q^{sam}, Q_{clip}, Q_{sam}), 
\end{aligned}
\end{equation}

To generate the final prediction, we merge the predictions from CLIP features and SAM features using a weighted linear combination:
\begin{equation}
\begin{aligned}
P = \alpha P_{sam} + ( 1 - \alpha) P_{clip}.
\end{aligned}
\end{equation}
where $\alpha$ represents the fusion weight that balances the contributions of the two predictions. This fusion strategy leverages the complementary strengths of both modalities, CLIP features provide rich semantic information from the text and visual domains, while SAM features offer precise spatial localization and segmentation capabilities. This approach not only enhances the robustness of the prediction but also improves the overall segmentation accuracy by combining the distinct advantages of both CLIP and SAM features.

After obtaining the final prediction, we employ the binary cross-entropy loss to supervise the segmentation process:
\begin{equation}\label{eq2}
\begin{aligned}
loss = loss_{ref} + \beta loss_{pred},
\end{aligned}
\end{equation}
where $loss_{ref}$ comes from the supervision of the refined local guidance in Eq.~\ref{eq1}, and $loss_{pred}$ is designed to supervise the final prediction by query label $M_q$:
\begin{equation}
\begin{aligned}
loss_{pred} = BCE(P, M_q).
\end{aligned}
\end{equation}

\begin{table*}[ht]
\caption{Performance comparisons with mIoU (\%) as a metric on PASCAL-5$^{i}$ dataset, the \textbf{bold} indicates the optimal performance. $*$ means the reproduce performance for the fair comparison.}
\label{tab:tab1}
\centering
\resizebox{\textwidth}{!}{$
\begin{tabular}{llccccc|ccccc}
\hline
\multirow{2}*{Method}  &\multicolumn{6}{c|}{1-shot} &\multicolumn{5}{c}{5-shot}\\ 
\cline{3-12}
& &Fold0 &Fold1 &Fold2 &Fold3 &Mean &Fold0 &Fold1 &Fold2 &Fold3 &Mean\\
\hline
PFENet (TPAMI'20)~\cite{tian2020prior} &  & 61.7 & 69.5 & 55.4 & 56.3 & 60.8 & 63.1 & 70.7 & 55.8 & 57.9 & 61.9 \\
SCL (CVPR'21)~\cite{zhang2021self} &  & 63.0 & 70.0 & 56.5 & 57.7 & 61.8 & 64.5 & 70.9 & 57.3 & 58.7 & 62.9 \\
PST (TNNLS'21)~\cite{yang2021part} &  & 52.2 & 67.6 & 54.4 & 52.4 & 56.6 & 56.5 & 68.3 & 55.6 & 53.2 & 58.4 \\
MSGA (TNNLS'22)~\cite{gao2022mutually} &  & 62.3 & 68.2 & 60.5 & 59.7 & 62.7 & 64.2 & 71.2 & 60.5 & 57.6 & 63.4s \\
SSP (ECCV'22)~\cite{fan2022self} &  & 60.5 & 67.8 & 66.4 & 51.0 & 61.4 & 67.5 & 72.3 & 75.2 & 62.1 & 69.3\\
DCAMA (ECCV'22)~\cite{shi2022dense} &  & 67.5 & 72.3 & 59.6 & 59.0 & 64.6 & 70.5 & 73.9 & 63.7 & 65.8 & 68.5\\
NERTNet (CVPR'22)~\cite{liu2022learning} &  & 65.4 & 72.3 & 59.4 & 59.8 & 64.2 & 66.2 & 72.8 & 61.7 & 62.2 & 65.7 \\
IPMT (NeurIPS'22)~\cite{liu2022intermediate} &  & 72.8 & 73.7 & 59.2 & 61.6 & 66.8 & 73.1 & 74.7 & 61.6 & 63.4 & 68.2 \\ 
ABCNet (CVPR'23)~\cite{wang2023rethinking} &  & 68.8 & 73.4 & 62.3 & 59.5 & 66.0 & 71.7 & 74.2 & 65.4 & 67.0 & 69.6 \\
MIANet (CVPR'23)~\cite{yang2023mianet} &  & 68.5 & 75.8 & 67.5 & 63.2 & 68.8 & 70.2 & 77.4 & 70.0 & 68.8 & 71.6 \\
HDMNet (CVPR'23)~\cite{peng2023hierarchical} &  & 71.0 & 75.4 & 68.9 & 62.1 & 69.4 & 71.3 & 76.2 & 71.3 & 68.5 & 71.8 \\
MSI (ICCV'23)~\cite{moon2023msi} &  & 71.0 & 72.5 & 63.8 & 65.9 & 68.3 & 73.0 & 74.2 & 66.6 & 70.5 & 71.1 \\
BAM (TPAMI'23)~\cite{lang2023base} &  & 69.9 & 75.4 & 67.0 & 62.1 & 68.6 & 72.6 & 77.1 & 70.7 & 69.8 & 72.6 \\
HPA (TPAMI'23)~\cite{cheng2022holistic} &  & 67.2 & 73.1 & 64.3 & 59.8 & 66.1 & 68.3 & 75.2 & 66.4 & 67.7 & 69.4 \\
PFENet++ (TPAMI'23)~\cite{luo2023pfenet++} &  & 63.3 & 71.0 & 65.9 & 59.6 & 64.9 & 66.1 & 75.0 & 74.1 & 64.3 & 69.9 \\
RiFeNet(AAAI'24)~\cite{bao2024relevant} &  & 68.9 & 73.8 & 66.2 & 60.3 & 67.3 & 70.4 & 74.5 & 68.3 & 63.4 & 69.2 \\
AENet(ECCV'24)~\cite{xu2024eliminating} &  & 71.3 & 75.9 & 68.6 & 65.4 & 70.3 & 73.9 & 77.8 & 73.3 & 72.0 & 74.2 \\
DCP (IJCV'24)~\cite{lang2024few} &  & 68.9 & 74.2 & 63.3 & 62.7 & 67.3 & 72.1 & 77.1 & 66.5 & 70.5 & 71.5 \\
CGMGM(AAAI'24)~\cite{shen2024cgmgm} &  & 71.1 & 75.0 & 69.6 & 63.7 & 69.9 & 71.8 & 78.9 & 69.1 & 68.6 & 72.0 \\
RD (CVPR'24)~\cite{zhou2024unlocking} &  & - & - & - & - & 77.7 & - & - & - & - & 78.0 \\
HMNet(NeurIPS'24)~\cite{xu2024hybrid} &  & 72.2 & 75.4 & 70.0 & 63.9 & 70.4 & 74.2 & 77.3 & 74.1 & 70.9 & 74.1 \\
LayerMI (TNNLS'24)~\cite{luo2024layer} &  & 63.6 & 74.6 & 67.4 & 66.6 & 68.1 & 72.3 & 77.0 & 75.6 & 70.8 & 73.9 \\
PAT (TPAMI'24)~\cite{bi2024prompt} &  & 68.3 & 73.2 & 66.2 & 60.1 & 67.0 & 73.3 & 77.6 & 75.1 & 69.5 & 73.9 \\
VRP-SAM$^{*}$ (CVPR'24)~\cite{sun2024vrp} &  & 73.9 & 78.3 & 70.6 & 65.0 & 71.9 & 76.3 & 76.8 & 69.5 & 63.1 & 71.4 \\
FCP (AAAI'25)~\cite{park2025foreground} &  & 74.9 & 77.4 & 71.8 & 68.8 & 73.2 & 77.2 & 78.8 & 72.2 & 67.7 & 74.0 \\
FBINet (TIM'25)~\cite{huang2025fbinet} &  & 67.4 & 71.7 & 63.1 & 63.1 & 66.3 & 69.2 & 75.1 & 66.9 & 66.7 & 69.4 \\
\hline
USD-\textbf{ours} &  & \textbf{79.8} & \textbf{82.9} & \textbf{76.6} & \textbf{74.1} &\textbf{78.4} &\textbf{80.2} &\textbf{83.1} &{\textbf{76.8}} & \textbf{74.3} &\textbf{78.6}  \\\hline
\end{tabular}%
$}
\end{table*}

\begin{table*}[!t]
\caption{Performance comparisons on COCO-20$^{i}$ dataset, the \textbf{bold} indicates the optimal performance. $*$ means the reproduce performance for the fair comparison.}
\label{tab:tab2}
\centering
\resizebox{\textwidth}{!}{$
\begin{tabular}{llccccc|ccccc}
\hline
\multirow{2}*{Method} & &\multicolumn{5}{c|}{1-shot} &\multicolumn{5}{c}{5-shot}\\ 
\cline{3-12}
& &Fold0 &Fold1 &Fold2 &Fold3 &Mean &Fold0 &Fold1 &Fold2 &Fold3 &Mean\\
\hline
PFENet (TPAMI'20)~\cite{tian2020prior} &  & 34.3 & 33.0 & 32.3 & 30.1 & 32.4 & 38.5 & 38.6 & 38.2 & 34.3 & 37.4 \\
SCL (CVPR'21)~\cite{zhang2021self} &  & 36.4 & 38.6 & 37.5 & 35.4 & 37.0 & 38.9 & 40.5 & 41.5 & 38.7 & 39.9 \\
PST (TNNLS'21)~\cite{yang2021part} &  & 30.4 & 37.5 & 30.2 & 32.6 & 32.7 & 34.0 & 42.2 & 34.7 & 39.0 & 37.5 \\
SSP (ECCV'22)~\cite{fan2022self} &  & 39.1 & 45.1 & 42.7 & 41.2 & 42.0 & 47.4 & 54.5 & 50.4 & 49.6 & 50.2\\
DCAMA (ECCV'22)~\cite{shi2022dense} &  & 41.9 & 45.1 & 44.4 & 41.7 & 43.3 & 45.9 & 50.5 & 50.7 & 46.0 & 48.3\\
NERTNet (CVPR'22)~\cite{liu2022learning} &  & 38.3 & 40.4 & 39.5 & 38.1 & 39.1 & 42.3 & 44.4 & 44.2 & 41.7 & 43.2 \\
IPMT (NeurIPS'22)~\cite{liu2022intermediate} &  & 41.4 & 45.1 & 45.6 & 40.0 & 43.0 & 43.5 & 49.7 & 48.7 & 47.9 & 47.5 \\ 
ABCNet (CVPR'23)~\cite{wang2023rethinking} &  & 42.3 & 46.2 & 46.0 & 42.0 & 44.1 & 45.5 & 51.7 & 52.6 & 46.4 & 49.1 \\
MIANet (CVPR'23)~\cite{yang2023mianet} &  & 42.5 & 53.0 & 47.8 & 47.4 & 47.7 & 45.9 & 58.2 & 51.3 & 52.0 & 51.7 \\
HDMNet (CVPR'23)~\cite{peng2023hierarchical} &  & 43.8 & 55.3 & 51.6 & 49.4 & 50.0 & 50.6 & 61.6 & 55.7 & 56.0 & 56.0 \\
MSI (ICCV'23)~\cite{moon2023msi} &  & 42.4 & 49.2 & 49.4 & 46.1 & 46.8 & 47.1 & 54.9 & 54.1 & 51.9 & 52.0 \\
BAM (TPAMI'23)~\cite{lang2023base} &  &45.2  &55.1  &48.7  &45.0  &48.5  &48.3  &58.4  &52.7  &51.4  &52.7  \\ 
HPA (TPAMI'23)~\cite{cheng2022holistic} &  & 43.2 & 50.5 & 45.5 & 46.2 & 46.3 & 49.4 & 58.4 & 52.5 & 50.9 & 52.8 \\
PFENet++ (TPAMI'23)~\cite{luo2023pfenet++} &  & 42.0 & 44.1 & 41.0 & 39.4 & 41.6 & 47.3 & 55.1 & 50.1 & 50.1 & 50.7 \\
RiFeNet(AAAI'24)~\cite{bao2024relevant} &  & 39.1 & 47.2 & 44.6 & 45.4 & 44.1 & 44.3 & 52.4 & 49.3 & 48.4 & 48.6 \\
AENet(ECCV'24)~\cite{xu2024eliminating} &  & 45.4 & 57.1 & 52.6 & 50.0 & 51.3 & 52.7 & 62.6 & 56.8 & 56.1 & 57.1 \\
DCP (IJCV'24)~\cite{lang2024few} &  & 43.0 & 48.6 & 45.4 & 44.8 & 45.5 & 47.0 & 54.7 & 51.7 & 50.0 & 50.9 \\
CGMGM(AAAI'24)~\cite{shen2024cgmgm} &  & 47.0 & 49.3 & 48.8 & 44.4 & 47.4 & 50.3 & 54.6 & 51.3 & 51.8 & 52.0 \\
RD (CVPR'24)~\cite{zhou2024unlocking} &  & - & - & - & - & 57.1 & - & - & - & - & 59.2 \\
HMNet(NeurIPS'24)~\cite{xu2024hybrid} &  & 45.5 & 58.7 & 52.9 & 51.4 & 52.1 & 53.4 & \textbf{64.6} & \textbf{60.8} & 56.8 & \textbf{58.9} \\
LayerMI (TNNLS'25)~\cite{luo2024layer} &  & 45.6 & 51.9 & 49.7 & 43.3 & 47.6 & 51.3 & 63.6 & 59.5 & 53.4 & 57.0 \\
PAT (TPAMI'24)~\cite{bi2024prompt} &  & 40.6 & 51.9 & 49.0 & 50.7 &48.0 & 51.2 & 63.2 & 59.5 & 55.9 &57.4 \\
VRP-SAM$^{*}$(CVPR'24)~\cite{sun2024vrp} &  & 44.3 & 54.3 & 52.3 & 50.0 & 50.2 & 50.5 & 59.5 & 56.9 &54.9 & 55.5 \\
FCP (AAAI'25)~\cite{park2025foreground} &  & 46.4 & 56.4 & 55.3 & 51.8 & 52.5 & 52.6 & 63.3 & 59.8 & 56.1 & 58.0 \\
FBINet (TIM'25)~\cite{huang2025fbinet} &  & 36.1 & 49.2 & 45.2 & 42.8 & 43.3 & 39.3 & 52.6 & 47.4 & 44.9 & 46.1 \\
\hline
USD-\textbf{ours} & & \textbf{52.6} & \textbf{63.5} & \textbf{58.4} &\textbf{57.0} &\textbf{57.9} & \textbf{53.8}  &64.0 &59.8 &\textbf{57.2} & 58.7 \\\hline
\end{tabular}%
$}
\end{table*}

\begin{figure}[!t]
	\centering
	\includegraphics[width=0.7\columnwidth]{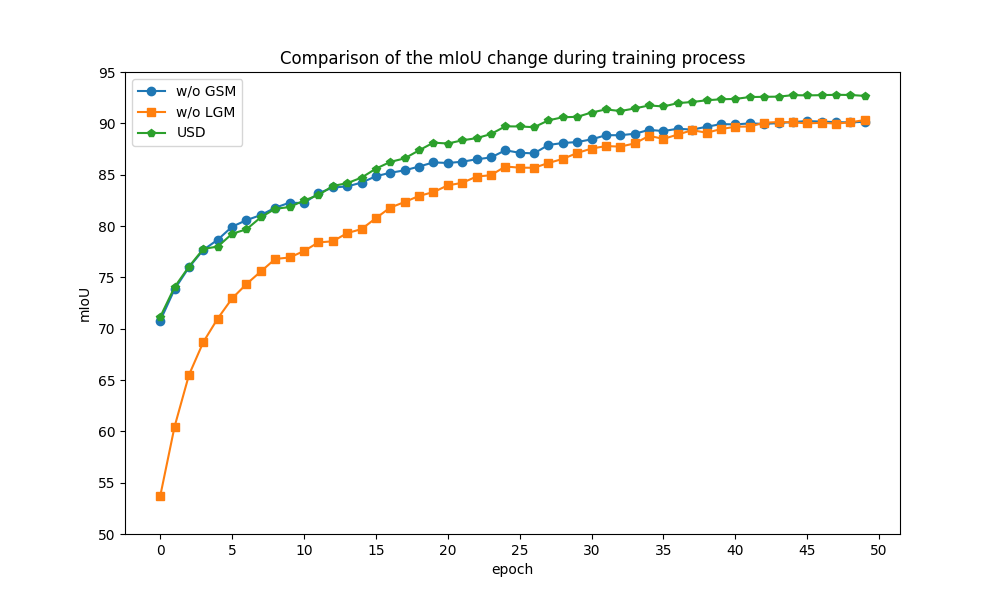}
        \vspace{-1.5em}
	\caption{Changes in mIoU at different experimental settings during the training process for PASCAL-5$i$ dataset, USD reaches its optimal effect and stabilizes before 50 epochs, faster than the previous method that required 100 epochs or more.}
	\label{fig:miou}
\end{figure}

\begin{figure}[!t]
	\centering
	\includegraphics[width=0.7\columnwidth]{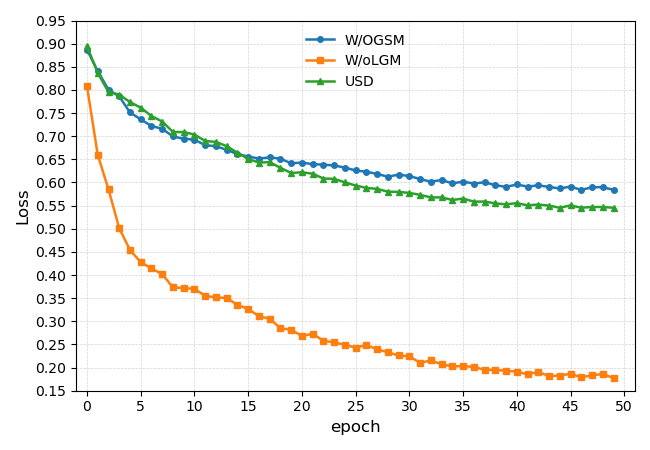}
        \vspace{-1.5em}
	\caption{Changes in Loss at different experimental settings during the training process for PASCAL-5$i$ dataset.}
	\label{fig:loss}
\end{figure}

\begin{algorithm}[!h]
\renewcommand{\algorithmicrequire}{\textbf{Input:}}
\renewcommand{\algorithmicensure}{\textbf{Output:}}
\caption{Unbiased Semantic Decoding (USD) for Few-shot Segmentation}
\label{alg:usd}
\begin{algorithmic}[1]
\REQUIRE Support image $I_s$ with mask $M_s$, query image $I_q$, target text prompt $t_f$, non-target text prompt $t_b$.
\ENSURE Segmentation mask prediction $P$ for query image $I_q$.

\textbf{Step 1: Encode visual and text features}
\STATE $F_s^{\text{clip}_v} \leftarrow E_{\text{clip}}^v(I_s)$, $F_q^{\text{clip}_v} \leftarrow E_{\text{clip}}^v(I_q)$,  $F_q^{\text{sam}} \leftarrow E_{\text{sam}}(I_q)$
\STATE $F_f^t \leftarrow E_{\text{clip}}^t(t_f)$, \quad $F_f^b \leftarrow E_{\text{clip}}^t(t_b)$

\textbf{Step 2: Global Supplement Module (GSM)}
\STATE $P_s \leftarrow \text{MAP}(\text{MLP}_1((F_s^{\text{clip}_v})), M_s)$,\\
$F_{\text{clip}}^q \leftarrow \text{MLP}_2((F_s^{\text{clip}_v}))$

\textbf{ Step 3: Local Guidance Module (LGM)}
\STATE $G_{\text{ini}} \leftarrow \text{SoftmaxGradCAM}(F_s^{\text{clip}_v}, t_f, t_b)$ 
\STATE $F_q^{\text{con}} \leftarrow \phi_1(\{P_s; F_q^{\text{clip}}; G_{\text{ini}})$, $F_q^{\text{self}}, H \leftarrow \text{SelfAttn}(F_{\text{con}}^q)$, $G_{\text{fin}} \leftarrow \max(H \odot G_{\text{ini}} \odot B, G_{\text{ini}})$
\STATE $F_q^{\text{fin}} \leftarrow \phi_2([F_q^{\text{self}}; G_{\text{fin}}])$ 

\textbf{ Step 4: Visual-Text Target Prompt Generator (VTPG)}
\STATE $Q_{\text{clip}} \leftarrow \text{CrossAttn}(t_f, F_q^{\text{fin}})$, $Q_{\text{sam}} \leftarrow \text{CrossAttn}(t_f, F_q^{\text{sam}})$

\textbf{ Step 5: Decode and fuse predictions}
\STATE $P_{\text{clip}}, P_{\text{sam}} \leftarrow D_{\text{sam}}(F_q^{\text{fin}}, F_q^{\text{sam}}, Q_{\text{clip}}, Q_{\text{sam}})$
\STATE $P \leftarrow \alpha P_{\text{sam}} + (1-\alpha) P_{\text{clip}}$ 

\STATE \textbf{return} $P$.
\end{algorithmic}
\end{algorithm}

\section{Experiments}
\subsection{Datasets and Evaluation Metrics.}
To assess the effectiveness of our proposed approach, we employed the PASCAL-5$^{i}$~\cite{shaban2017one} and COCO-20$^{i}$~\cite{nguyen2019feature} datasets. The PASCAL-5$^{i}$~\cite{shaban2017one} is derived from the PASCAL VOC 2012 dataset~\cite{everingham2010pascal} and enhanced with SDS~\cite{hariharan2011semantic}. It is a well-established dataset for segmentation tasks, featuring 20 distinct object classes like people, cars, cats, dogs, chairs, airplanes, and others. The COCO-20$^{i}$~\cite{nguyen2019feature} dataset, based on MSCOCO~\cite{lin2014microsoft}, comprises over 120,000 images spanning 80 categories and presents a more complex challenge.
For the fair evaluation of our proposed method, we utilized the mean intersection-over-union (mIoU) and foreground-background IoU (FB-IoU) as the evaluation metrics, in line with prior research~\cite{lang2022learning, peng2023hierarchical, tian2020prior}.

\subsection{Implementation details.}
In all experiments on PASCAL-5$^{i}$ and COCO-20$^{i}$, the images are set to 512$\times$512 pixels, the CLIP pre-trained model is ViT-B-16~\cite{radford2021learning} and the SAM pre-trained model is SAM-h~\cite{kirillov2023segment}. For the 5-shot case, the final segmentation mask is generated by aggregating the predictions from five different support sets. As shown in Fig.~\ref{fig:miou}, under the guidance of semantic abundant feature expression ability and sufficient target information, our proposed USD method can achieve better performance and stabilization more rapidly. Besides, to verify the convergence of our model, we plot its loss curve in Fig.~\ref{fig:loss}. Without LGM, USD converges more slowly; without GSM, the convergence curve shows larger fluctuations. This demonstrates that fusing robust CLIP semantics with consistent SAM features enables more efficient target recognition, thereby accelerating model reasoning. All experiments on PASCAL-5$^{i}$ dataset and COCO-20$^{i}$ dataset only require 50 epochs of training, compared to 100 epochs or more for prior methods, this significantly reduces the training time, and we only utilize two GeForce RTX 3090s to train the USD model. The batch size is set to 8, and the learning rate is set to $4\times10^{-4}$, $\alpha=\beta=0.5$. Other settings such as image enhancement methods, optimizers, etc., are the same as before~\cite{sun2024vrp}. Besides, in order to more accurately distinguish between target and non-target areas, we randomly selected 50 sample points from the final local guidance $G_{fin}$, and it contains 25 foreground points and 25 background points. Since the SAM-based FSS method VRP-SAM~\cite{sun2024vrp} did not perform 5-shot experiments, to be fair, we reproduce all the experimental results of VRP-SAM.

\textbf{Algorithm process and analyzing of the model.}
Algorithm~\ref{alg:usd} presents the procedural flow of our model. Leveraging strong text-visual alignment capability of CLIP, GSM and LGM are proposed to enhance semantic decoding of SAM and further alleviate the bias to base classes.

\begin{figure*}[ht]
	\centering
	\includegraphics[width=\textwidth]{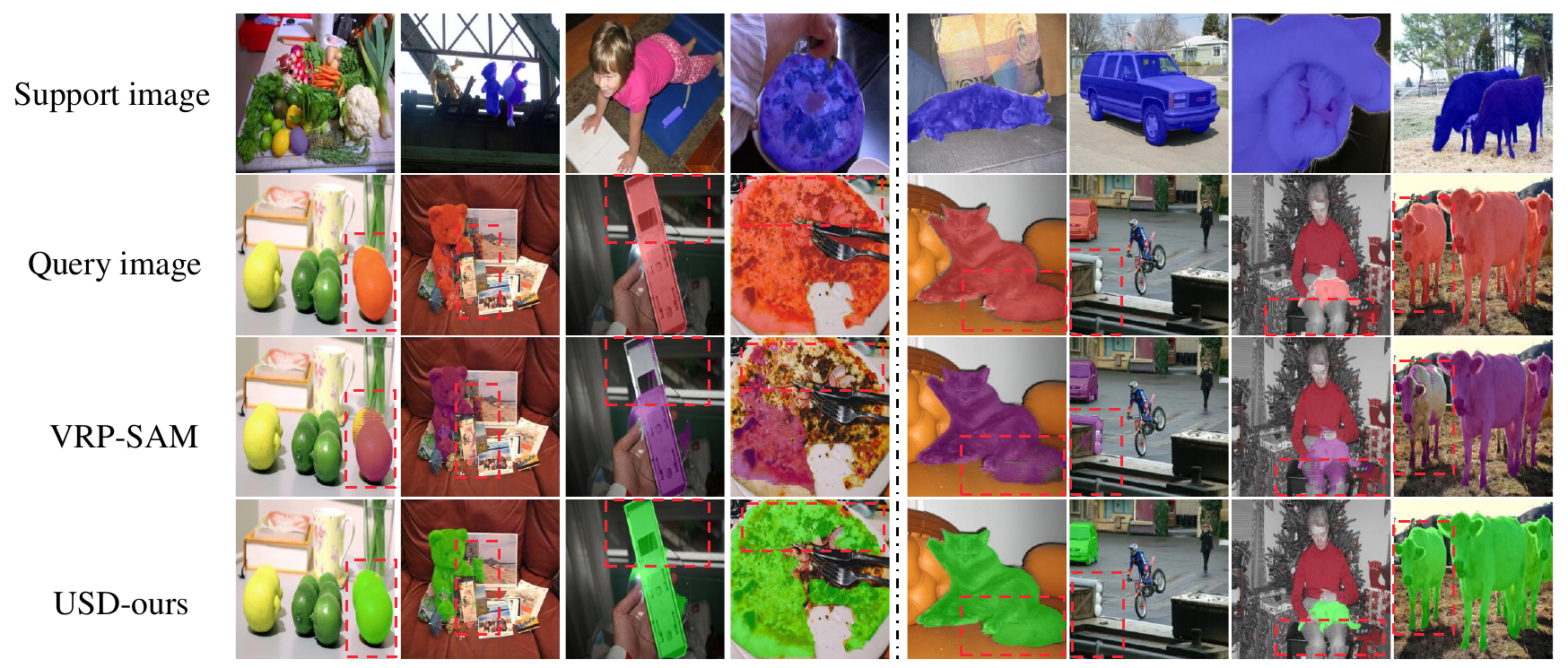}
	\caption{Qualitative results of the proposed USD and VRP-SAM approach under 1-shot setting from both PASCAL-5$^{i}$and COCO-20$^{i}$ datasets. Each row from top to bottom represents the support images with ground-truth (GT) masks (blue), query images with GT masks (red), VRP-SAM results (purple), and our results (green), respectively.}
	\label{fig:3}
\end{figure*}

\begin{table*}
\centering
\caption{Efficiency comparison of different methods and the proposed modules}
\label{tab_effeiciency}
\resizebox{\textwidth}{!}{%
\begin{tabular}{lcccccc}
\hline
\multirow{2}{*}{Method} & \multirow{2}{*}{GPUs} & \multirow{2}{*}{Inf Time} & \multirow{2}{*}{FLOPs} & \multirow{2}{*}{Learnable Params} & \multirow{2}{*}{mIoU} & \multirow{2}{*}{FB-IoU}\\
& & & & & &  \\
\hline
VRP-SAM (baseline)   &19.2GB  & 0.16s  & 889G & 1.6M    & 73.9 &84.6      \\
USD (ours) & 20.0GB  & 0.15s  & 543G & 11.9M & \textbf{79.8} & \textbf{89.2} \\
USD \emph{w/o} GSM & 18.6GB  & 0.10s  & 542G &11.6M  & 68.5  &80.3 \\
USD \emph{w/o} LGM & 11.6GB  & 0.12s  & 505G & 1.3M & 70.2 & 80.4 \\
\hline
\end{tabular}
}
\end{table*}

\subsection{Comparison with state-of-the-art}
\subsubsection{Quantitative results}
Table~\ref{tab:tab1} presents the performance of our method and existing state-of-the-art methods on the PASCAL-5$^{i}$ in few-shot segmentation task. Our approach significantly outperforms other methods on both 1-shot and 5-shot tasks, achieving new state-of-the-art results. Specifically, it improves the mIoU by 6.5\% over VRP-SAM~\cite{sun2024vrp} and by 5.2\% over FCP~\cite{park2025foreground} on 1-shot tasks, both of which are SAM-based methods. For the 5-shot segmentation task, our approach outperforms other approaches by a clear margin, with mIoU gain of 7.2$\%$ for VRP-SAM~\cite{sun2024vrp} and 4.6$\%$  FCP~\cite{park2025foreground}, respectively. 

In Table~\ref{tab:tab2}, we compare the performance of our approach and others on COCO-20$^{i}$ dataset. Our approach also exhibits strong performance and achieves new
state-of-the-art performance. Specifically, our approach improves the VRP-SAM by 7.7$\%$ and 3.2$\%$ mIoU for 1-shot and 5-shot tasks, compared to FCP~\cite{park2025foreground}, our method raises 5.4$\%$ at 1-shot and 0.7$\%$ at 5-shot.

\subsubsection{Qualitative results}
In order to better show the effectiveness of our proposed model, we visualize the results of another SAM-based method VRP-SAM and our proposed method in Fig.~\ref{fig:3}, it can be found that our method (green part) has a much stronger target localization ability than the VRP-SAM (purple part), and the consistency of the segmentation result is stronger, even for unlabelled regions in the human annotation, our method is able to obtain a more complete segmentation through feature consistency, such as the unlabelled bear within the image in the second column and the poorly labeled cat in the penultimate column.

\begin{figure*}[!t]
	\centering
	\includegraphics[width=\textwidth]{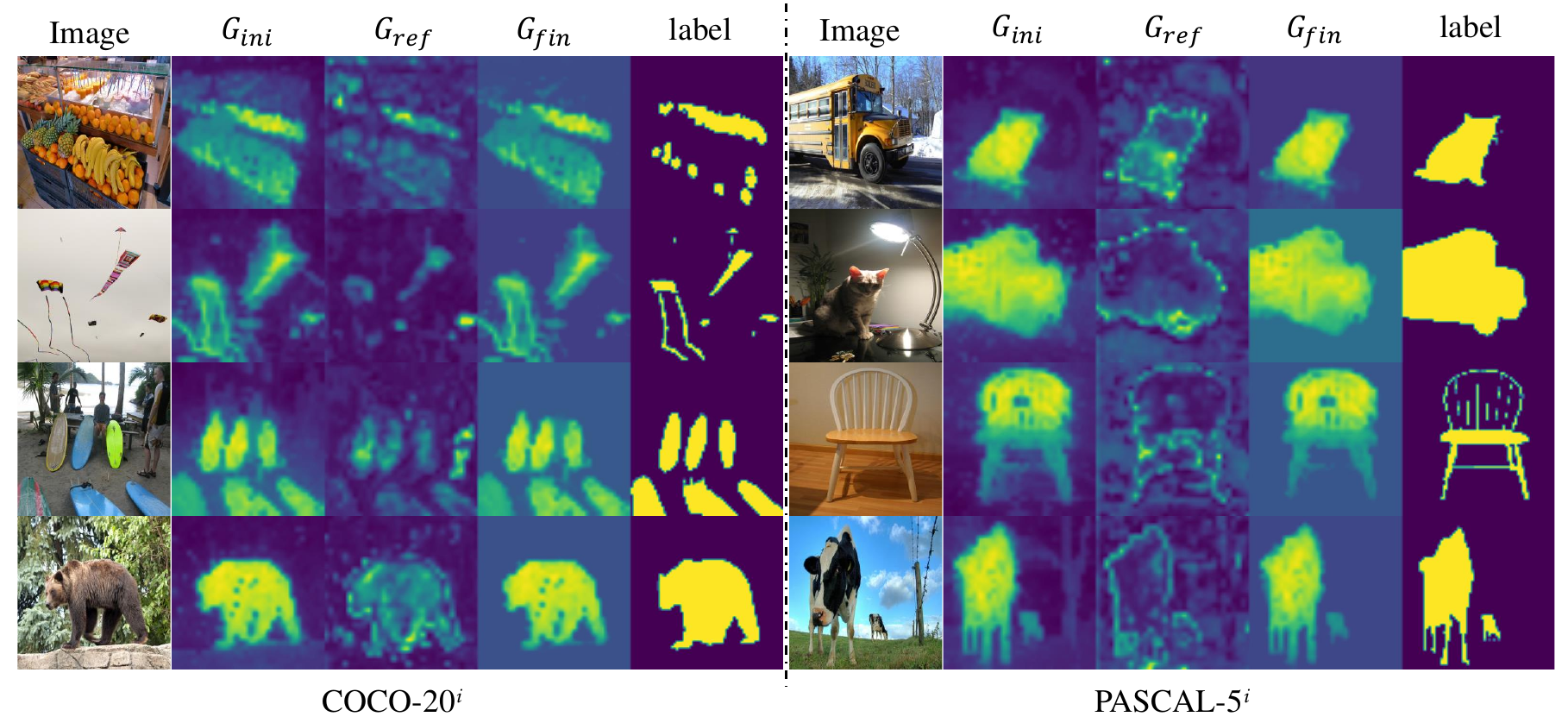}
	\caption{Visualization of the different prior information generated by our proposed method. The left is sampled from  COCO-20$^{i}$~\cite{nguyen2019feature} and the right is selected from PASCAL-5$^{i}$~\cite{shaban2017one}. Each column in each dataset ``Image" represents the query image, $G_{ini}$ represents the initial guidance, $G_{ref}$ represents the guidance after the dynamic refinement, $G_{fin}$ means the final guidance.  $G_{ini}$ focus more on the central area of the target and $G_{ref}$ expand more target areas that $G_{ini}$ don not focused on. After the process of $G_{ini}$ and $G_{ref}$, $G_{fin}$ has stronger foreground consistency versus background consistency, which is more valuable for consistent and unbiased segmentation results.}
	\label{fig:cam}
\end{figure*}

Fig.~\ref{fig:cam} shows the visualization of our proposed local guidance $G_{ini}$ and $G_{ref}$, aiming to help understand the useful localization capabilities of $G_{ini}$ and the ability to expand focus on the target area of $G_{ref}$. $G_{ini}$ focuses more on the accurate target regions, which are localized in a limited region compared to the whole object. $G_{ref}$, on the other hand, focuses on target areas not attended to by $G_{ini}$, by complementing the two, more useful target information is extracted. Moreover, the internal consistency of target features and the non-target features in the final local guidance $G_{fin}$ have been strengthened, and the internal consistency of features can help the model better adapt to the distribution of different categories of data and enhance the generalization ability of the few-shot model.

\begin{figure}[t]
	\centering
	\includegraphics[width=0.7\columnwidth]{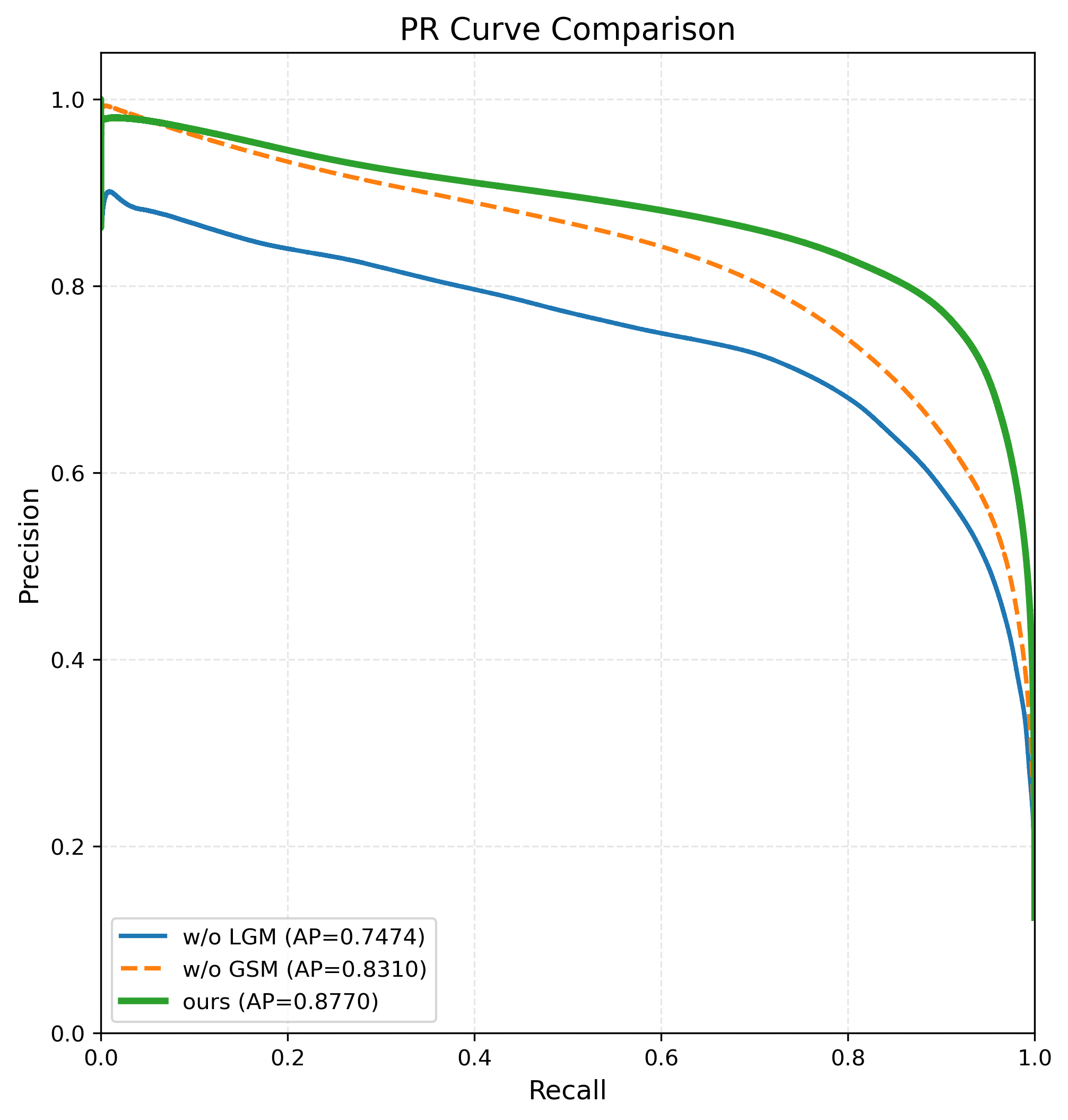}
        \vspace{-1.5em}
	\caption{Precision-Recall Curve performance under different settings.}
	\label{fig:pr}
\end{figure}

Fig.~\ref{fig:pr} shows the Precision-Recall curve on our proposed method and the designed modules, the method we proposed (AP=0.877) is superior to the control models without LGM (AP=0.747) and GSM (AP=0.831). The overall curve is further to the upper right, verifying that both the LGM and GSM modules are effective in reducing bias to base class and improving the overall detection performance, and the contribution of LGM is more useful.

\begin{table*}[ht]
\caption{Domain-shift performance with mIoU (\%) as a metric on COCO-20$^{i}$ dataset to PASCAL-5$^{i}$ dataset, the \textbf{bold} indicates the optimal performance.}
\label{tab_shift}
\centering
\resizebox{\textwidth}{!}{$
\begin{tabular}{llccccc|ccccc}
\hline
\multirow{2}*{Method}  &\multicolumn{6}{c|}{1-shot} &\multicolumn{5}{c}{5-shot}\\ 
\cline{3-12}
& &Fold0 &Fold1 &Fold2 &Fold3 &Mean &Fold0 &Fold1 &Fold2 &Fold3 &Mean\\
\hline
PFENet (TPAMI'20)~\cite{tian2020prior} &  & 43.2 & 65.1 & 66.5 & 69.7 & 61.1 & 45.1 & 66.8 & 68.5 & 73.1 & 63.4 \\
RePRI (CVPR'21)~\cite{sun2024vrp} &  & 52.2 & 64.3 & 64.8 & 71.6 & 63.2 & 56.5 & 68.2 & 70.0 & 76.2 & 67.7 \\
\hline
USD-\textbf{ours} &  & \textbf{68.0} & \textbf{78.7} & \textbf{79.3} & \textbf{89.5} &\textbf{78.8} &\textbf{68.1} &\textbf{79.0} &{\textbf{79.6}} & \textbf{89.7} &\textbf{79.1}  \\\hline
\end{tabular}%
$}
\end{table*}

\subsubsection{Efficiency Comparison}
As shown in Table~\ref{tab_effeiciency}, our method achieves a 5.9$\%$ mIoU improvement over the baseline while significantly reducing computational complexity and maintaining minimal memory overhead. This performance gain is attributed to the GSM module, which effectively captures global semantic relationships, and the LGM module, which provides pixel-level local guidance. Notably, the LGM module contains more learnable parameters due to its design incorporating a learnable transformer that dynamically fine-tunes the prior representation. This dynamic fine-tuning mechanism is crucial, as it enables the model to rapidly adapt to novel classes.

\subsubsection{Domain-shift and low-resource settings.}
As shown in Table~\ref{tab_shift}, we conduct cross-domain experiments from COCO-20$i$ to PASCAL-5$i$ following the setting of RePRI[69]. The experimental results show that even on datasets with different data distributions, our proposed method can maintain very impressive performance. In addition, as shown in Table~\ref{tab_low_source}, we also verify the performance of our proposed method in a more challenging experimental setting. Specifically, we reduce the training data to 10$\%$ of the original size while keeping the number of test images unchanged. The results show that our method still maintains strong performance under low-resource conditions.

\begin{table}[!t]
\caption{Results of USD under low source settings on the PASCAL-5$^{i}$ dataset.}
\centering
\normalsize
\begin{tabular}{c|cc|cc}
\hline
\multicolumn{5}{c}{\textbf{USD}} \\
\hline
& \multicolumn{2}{c|}{\textbf{1-shot}} & \multicolumn{2}{c}{\textbf{5-shot}} \\
\cline{2-5}
& mIoU ($\%$) & FB-IoU ($\%$) & mIoU ($\%$) & FB-IoU ($\%$) \\
\hline
Fold0 &75.5  &86.4  & 75.9 & 86.4 \\
\hline
Fold1 &77.8  &86.7  &78.1  & 86.9 \\
\hline
Fold2 &73.4  &81.2  &73.7  & 81.6 \\
\hline
Fold3 &69.7  &81.1  &69.8  & 81.5 \\
\hline
Mean & 74.1 &83.6  &74.6  &84.1  \\
\hline
\end{tabular}
\label{tab_low_source}
\end{table}

\begin{figure}[!t]
\centering
\begin{tikzpicture}[scale=0.6]
\begin{axis}[
    title={Ablation Study on Loss Percentage},
    xlabel={$\beta$},
    ylabel={Performance (\%)},
    xmin=0, xmax=1,
    xtick={0, 0.2, 0.5, 0.8, 1},
    ymin=75, ymax=90,
    legend style={
    draw=black,
    fill=white, 
    fill opacity=0.9,
    draw opacity=1,
    text opacity=1, 
    at={(0.05,0.5)}, 
    anchor=west,
    legend cell align=left
},
    grid style=dashed,
]
\addplot[
    color=blue,
    mark=square,
    ]
    coordinates {
    (0,78.18)(0.2,77.30)(0.5,79.80)(0.8,79.69)(1,78.39)
    };
    \addlegendentry{mIoU}
\addplot[
    color=red,
    mark=triangle,
    ]
    coordinates {
    (0,87.80)(0.2,85.15)(0.5,89.28)(0.8,89.13)(1,87.93)
    };
    \addlegendentry{FB-IoU}

\end{axis}
\end{tikzpicture}
\caption{Ablation study on loss percentage. $\beta$ denotes the weights in the fusion process of the refinement loss $loss_{ref}$ and final loss $loss_{fin}$.}
\label{fig:loss_ablation}
\end{figure}
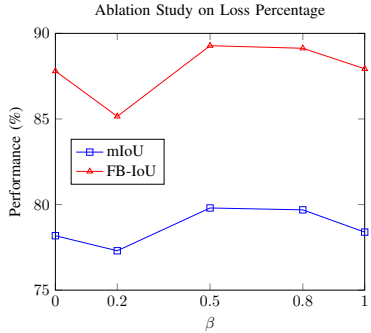

\subsection{Ablation Study}
To verify the usefulness of our proposed module, we conduct a series of ablation studies of each module on the PASCAL-5$^{i}$ dataset.
\subsubsection{Ablation Study on GSM and LGM}
As shown in Table~\ref{tab_module}, removing the proposed Global Supplement Module (GSM) and Local Semantic Module (LGM) results in a model performance decrease of 1.58$\%$, and 11.57$\%$, respectively. This indicates that both global semantic guidance and local semantic guidance are crucial for the consistent segmentation of SAM. Specifically, global semantic guidance helps the model capture overall contextual information, which is essential for distinguishing objects from the background. Meanwhile, local semantic guidance provides fine-grained pixel-level information, enabling the model to accurately segment objects with complex shapes and boundaries.

\begin{table}[!t]
\caption{Ablation study about our proposed GSM and LGM on the PASCAL-5$^{i}$, ``USD" represents the proposed method, ``GSM" and ``LGM" represent the proposed GSM module and LGM module.}
\centering
\normalsize 
\begin{tabular}{ccc}
\hline
&mIoU ($\%$) &FB-IoU ($\%$) \\
\hline
USD  &\textbf{79.80} &\textbf{89.28} \\
\emph{w/o} GSM  &78.22 &87.18\\
\emph{w/o} LGM  &68.23 &79.74\\
\hline
\label{tab_module}
\end{tabular}
\end{table}

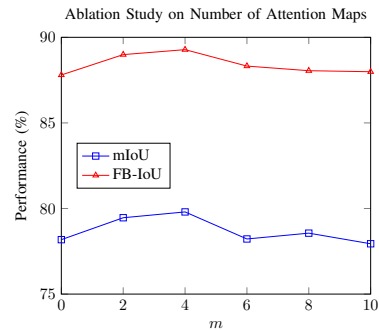
\begin{figure}[!t]
\centering
\begin{tikzpicture}[scale=0.6]
\begin{axis}[
    title={Ablation Study on Number of Attention Maps},
    xlabel={\( m \)},
    ylabel={Performance (\%)},
    xmin=0, xmax=10,
    xtick={0, 2, 4, 6, 8, 10},
    ymin=75, ymax=90,
    legend style={
    draw=black, 
    fill=white, 
    fill opacity=0.9,
    draw opacity=1,
    text opacity=1,
    at={(0.05,0.5)}, 
    anchor=west, 
    legend cell align=left 
},
    grid style=dashed,
]

\addplot[
    color=blue,
    mark=square,
    ]
    coordinates {
    (0,78.18)(2,79.46)(4,79.80)(6,78.22)(8,78.56)(10,77.94)
    };
    \addlegendentry{mIoU}

\addplot[
    color=red,
    mark=triangle,
    ]
    coordinates {
    (0,87.80)(2,88.99)(4,89.28)(6,88.32)(8,88.05)(10,87.99)
    };
    \addlegendentry{FB-IoU}

\end{axis}
\end{tikzpicture}
\caption{Ablation study on the number of attention maps \( m \) for refinement.}
\label{fig:attention_maps_ablation}
\end{figure}

 \begin{table}[!t]
 \caption{Ablation study about features sent to the decoder on the PASCAL-5$^{i}$, ``USD" represents the proposed method, ``CLIP" represents only the CLIP features are sent to the decoder and ``SAM" means only the SAM features are sent to the decoder.}
\centering
\normalsize  
\arrayrulecolor{black} 
\begin{tabular}{ccc}
\hline
&mIoU ($\%$) &FB-IoU ($\%$) \\
\hline
USD  &\textbf{79.80} &\textbf{89.28} \\
SAM  &76.82 &87.08\\
CLIP  &71.99 &84.55\\
\hline
\label{tab_fts}
\end{tabular}
\end{table}

 \subsubsection{Ablation Study on features sent to decoder}
As shown in table~\ref{tab_fts}, when only SAM features are utilized for decoding, the performance of the model is 76.82$\%$, which reduces 2.98$\%$. This indicates that while SAM features provide strong spatial and contextual information, they may lack sufficient semantic richness to fully capture the diversity of objects. When only CLIP features are utilized for decoding, the performance of the model is 71.99$\%$. Despite using only CLIP features in the frozen SAM decoder, the model still achieves a reasonably good performance. This further indicates that the GSM and LGM module effectively maps CLIP features into the SAM feature space. CLIP features, derived from a language-vision model, provide rich semantic information that can compensate for the semantic limitations of SAM features. The best performance is achieved only when the two features are fused, demonstrating the necessity of integrating CLIP and SAM features. The use of CLIP features supplements the semantic information lacking in SAM features, thereby producing superior segmentation results. This integration allows the model to leverage both the feature consistency of SAM and the rich semantics of CLIP features, leading to improved segmentation performance.

\begin{table}[t]
\caption{Ablation study about text prompts on PASCAL-5$^{i}$, $t_f$ represents the target prompt and $t_b$ represents the non-target prompt.}
\centering
\normalsize  
\resizebox{\columnwidth}{!}{$
\begin{tabular}{ccc}
\hline
                $t_f/t_b$              &mIoU ($\%$)            &FB-IoU ($\%$) \\\hline
``a photo of \{\emph{target class}\}"  & \multirow{2}{*}{\textbf{79.80}}   & \multirow{2}{*}{\textbf{89.28}}\\
``a photo without \{\emph{target class}\}"   \\  \hline
``a photo of \{\emph{target class}\}"  & \multirow{2}{*}{69.43}    & \multirow{2}{*}{82.24}\\
``not a photo of \{\emph{target class}\}"   \\  \hline
``a photo of \{\emph{target class}\}"  & \multirow{2}{*}{77.61}    & \multirow{2}{*}{88.20}\\
``a photo excluding \{\emph{target class}\}"   \\  \hline
``a snapshot of \{\emph{target class}\}"  & \multirow{2}{*}{70.45}    & \multirow{2}{*}{83.41}\\
``a snapshot without \{\emph{target class}\}"   \\  \hline
``a snapshot of \{\emph{target class}\}"  & \multirow{2}{*}{66.67}    & \multirow{2}{*}{80.03}\\
``a snapshot with no occurrence of \{\emph{target class}\}"   \\  \hline
``a capture of \{\emph{target class}\}"  & \multirow{2}{*}{75.24}    & \multirow{2}{*}{86.30}\\
``a capture devoid of \{\emph{target class}\}"   \\  \hline
\label{tab:text}
\end{tabular}
$}
\end{table}

\begin{table}[t]
\caption{Ablation Study on prediction percentage $\alpha$}
\centering
\normalsize  
\arrayrulecolor{black} 
\begin{tabular}{cccccc}
\hline
$\alpha$  &0 &0.3 & 0.5 &0.8 &1\\
\hline
mIoU ($\%$) &71.99 &79.06 &\textbf{79.80} &78.06 &76.82 \\
FB-IoU ($\%$) &84.55 &88.68  &\textbf{89.28} &87.94 &87.08 \\
\hline
\label{tab_pred}
\end{tabular}
\end{table}

\subsubsection{Ablation Study on loss percentage $\beta$}
The Fig.~\ref{fig:loss_ablation} demonstrates the experimental results at different loss ratios, and the experimental results show that the best experimental results are achieved only when $\beta=0.5$, which reaches 79.80$\%$ on mIoU.

\subsubsection{Ablation Study on amount of attention maps for refinement}
Since in LGM we use learnable attention maps to fine-tune the fixed local guidance to focus on a wider target area, we conduct ablation studies on the number of attention maps used for fine-tuning as shown in Fig.~\ref{fig:attention_maps_ablation}, and the experiments found that the model yielded the best results when the last four layers were taken, reaching 79.80$\%$ on mIoU. From the experiment we can see that if too much shallow information is utilized, for example, when using the average value from the last 10 or 8 layers to refine the fixed local guidance, the results of the experiment are rather worse, this is because, the shallow features of the CLIP focus more on the detail information, such as texture, color and so on, but this is detrimental to the construction of the semantic relationship between the pixels. When the deeper features are utilized more, such as the last 2 layers or the last 4 layers, the results of the model become significantly better, which also proves that the higher-level features of CLIP are more capable of extracting abstract semantic features, and that the higher-level features are helpful in constructing the semantic relationship between pixels, but only utilizing the higher-level features may be too global and lack of local detail information. The averaging of the last four layers can better balance the global semantics and local details, thus optimizing the model performance.

\subsubsection{Ablation Study on text prompt}
\begin{figure}[!t]
	\centering
	\includegraphics[width=1\columnwidth]{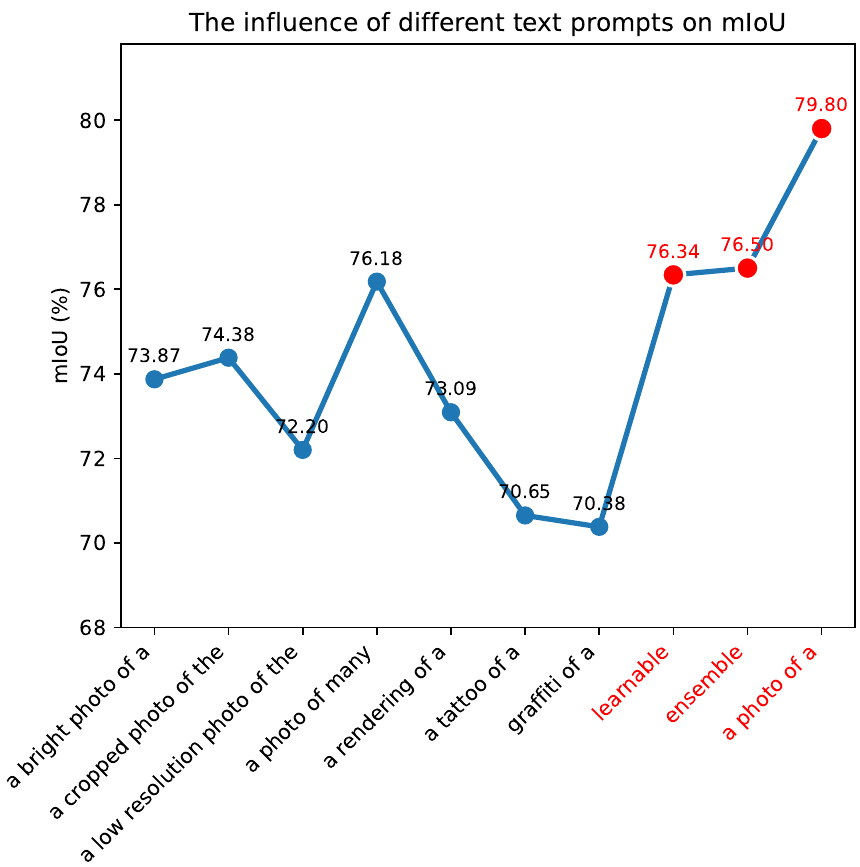}
        \vspace{-1.5em}
	\caption{Model sensitivity on more text prompts. ``learnable'' indicates that fine-tune the text prompt embeddings, and ``ensemble'' indicates that fuse all text prompt embeddings in the table. The figure only shows the representation of the foreground text. For the corresponding background text, we have changed ``of'' to ``without''}
	\label{text_miou}
\end{figure}

To investigate the impact of different text prompts on model performance, we designed various combinations of text prompts, where \(t_f\) represents the target prompt and \(t_b\) represents the non-target prompt. As shown in Table~\ref{tab:text}, when \(t_f\) is ``a photo of \{\emph{target class}\}" and \(t_b\) is ``a photo without \{\emph{target class}\}", the model achieves the best performance, with mIoU and FB-IoU reaching 79.80$\%$ and 89.28$\%$, respectively. Additionally, we further explore the sensitivity of our model to text descriptions. As shown in Fig.~\ref{text_miou}, the sensitivity to the text is verified through the following three aspects: (1) more prompt sets used in CLIP\footnote{\url{https://github.com/openai/CLIP/blob/main/notebooks}} are expanded to enrich the text representation; (2) a learnable text encoding module is introduced to dynamically optimize the representation of text features; (3) adopt multi-text fusion to integrate all text description information. However, different representation methods all experience a certain degree of performance degradation. It also proves that ``a photo of a  \{\emph{target class}\}'' and ``a photo without \{\emph{target class}\}'' is the most suitable text representation for our method.

\subsubsection{Ablation Study on prediction percentage $\alpha$}
Table.~\ref{tab_pred} demonstrates the experimental results at different prediction ratios, and the experimental results show that the best experimental results are achieved only when $\alpha=0.5$, which reaches 79.80$\%$ on mIoU and 89.28$\%$ on FB-IoU. This finding further indicates that the semantic features of CLIP and consistency of SAM can work well together. The strength of CLIP lies in understanding the meanings and contexts within images and text. The advantage of SAM is maintaining consistent and stable predictions across different parts of an image. When combined, these two models can leverage each other's strengths. By balancing the contributions of both features, the model benefits from the semantic richness provided by CLIP and the spatial consistency ensured by SAM. This synergistic combination allows the model to achieve an unbiased segmentation results.


\begin{table}[!t]
\centering
\normalsize  
\caption{Ablation study about extracts prompts from pure vision modal (\emph{w/o} $VTPG$) or extracts target information from both text and vision modal (\emph{w} $VTPG$) on the PASCAL-5$^{i}$.}
\label{tab_vtpg}
\begin{tabular}{ccc}
\hline 
&mIoU ($\%$) &FB-IoU ($\%$) \\
\hline
\emph{w/o} $VTPG$  &74.49 &85.96\\
\emph{w} $VTPG$  &\textbf{79.80} &\textbf{89.28} \\
\hline
\end{tabular}
\end{table}

\subsubsection{Ablation Study on VTPG}
As shown in the table~\ref{tab_vtpg}, in order to verify the effectiveness of the VTPG, we designe relevant ablation experiments, and find that when the VTPG is not used, \emph{i.e.,} when only visual information is utilized to extract the prompt, the mIoU of the model decreases to 74.49$\%$, which proving the effectiveness of the VTPG.

\subsubsection{Ablation Study on prompts source}
As shown in the table~\ref{tab_source}, our proposed USD simultaneously extracts valid target information from both the support set and the query set, which yields a more competitive performance due to the consistency of the information in the query image itself, enhancing the ability to generalize to new classes.

\begin{figure}[!t]
	\centering
	\includegraphics[width=1.0\columnwidth]{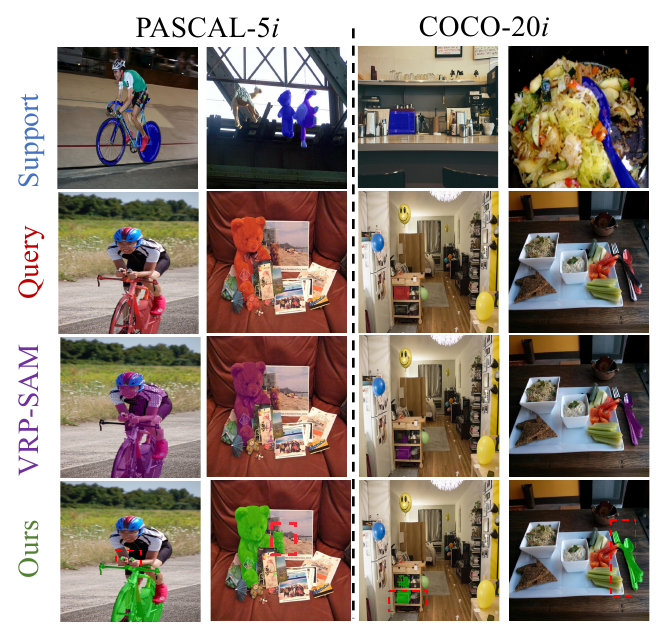}
        \vspace{-1.5em}
	\caption{Some failure cases on the PASCAL-5$i$ dataset and the COCO-20$i$ dataset, which are marked with red dashed boxes.}
	\label{table_failure}
\end{figure}

\begin{table}[t]
\centering
\caption{Ablation study about extracts prompts from support set ($S$) or extracts target information from both support and query sets ($S\&Q$) on the PASCAL-5$^{i}$.}
\centering
\normalsize   
\begin{tabular}{ccc}
\hline
&mIoU ($\%$) &FB-IoU ($\%$) \\
\hline
S  &72.16 &82.55\\
S\&Q  &\textbf{79.80} &\textbf{89.28} \\
\hline
\end{tabular}
\label{tab_source}
\end{table}

\subsubsection{Ablation Study on backbone settings}
To evaluate how backbone configurations affect model performance, we conduct ablation experiments on different variants of CLIP and SAM. As shown in Table~\ref{tab_backbone}, reducing the size of these pre-trained models leads to significant performance degradation. For SAM, smaller versions with fewer Transformer layers and narrower attention heads exhibit weaker feature extraction and long-range context modeling capabilities. Similarly, CLIP encoders with larger patch sizes lose fine-grained visual information, impairing their performance on tasks requiring detailed visual understanding.

Finally, as shown in Fig.~\ref{table_failure}, the proposed model may exhibit over-segmentation issues, primarily arising from the visual feature consistency inherent in SAM. This phenomenon manifests in two key aspects: (1) the erroneous segmentation of non-target objects with similar visual characteristics (e.g., table knives and forks), and (2) the segmentation of unannotated regions, such as the smaller bear photo in the second column.

\section{Conclusion}
This paper introduces the Unbiased Semantic Decoding (USD) framework, which enhances few-shot segmentation by integrating the vision foundation models. Different from the previous method which only extracts prompts from support samples, USD simultaneously extracts target information from both support and query sets for consistent predictions. The framework includes two feature enhancement strategies, GSM supplements SAM features from image-level through feature alignment, and LGM provides intensive semantic classification indicating for SAM features through semantic affinity. Based on the need for accurate prompt under SAM framework, VTPG creates target-focused embeddings by interacting text embeddings with visual features. Without retraining the vision foundation models, the proposed method achieves state-of-the-art performance. In the future, overcoming the excessive segmentation of the foreground region by the model remains a challenge worth investigating. This issue may be addressed by introducing more robust textual prompts or designing methods to suppress the activation of non-target foreground areas.

\begin{table}[!t]
\centering
\caption{Ablation study about the different SAM and CLIP backbone settings on the PASCAL-5$^{i}$.}
\normalsize 
\begin{tabular}{@{}llll@{}}
\toprule
\multicolumn{1}{c}{SAM} & \multicolumn{1}{c}{CLIP} & \multicolumn{1}{c}{mIoU} & \multicolumn{1}{c}{FB-IoU} \\ \midrule
SAM-H & Vit-B-16  & \textbf{79.80}  & \textbf{89.28} \\ \midrule
SAM-B & Vit-B-16  & 67.23   & 81.48   \\ \midrule
SAM-H & Vit-B-32  & 73.70  & 85.23  \\ \midrule
SAM-B & Vit-B-32  & 57.95 & 74.69  \\ \bottomrule
\end{tabular}
\label{tab_backbone}
\end{table}
{
    \small
    \bibliographystyle{IEEEtran}
    \bibliography{main}
}
\vspace{-4em}
\begin{IEEEbiography}[{\includegraphics[width=1in,height=1.25in,clip,keepaspectratio]{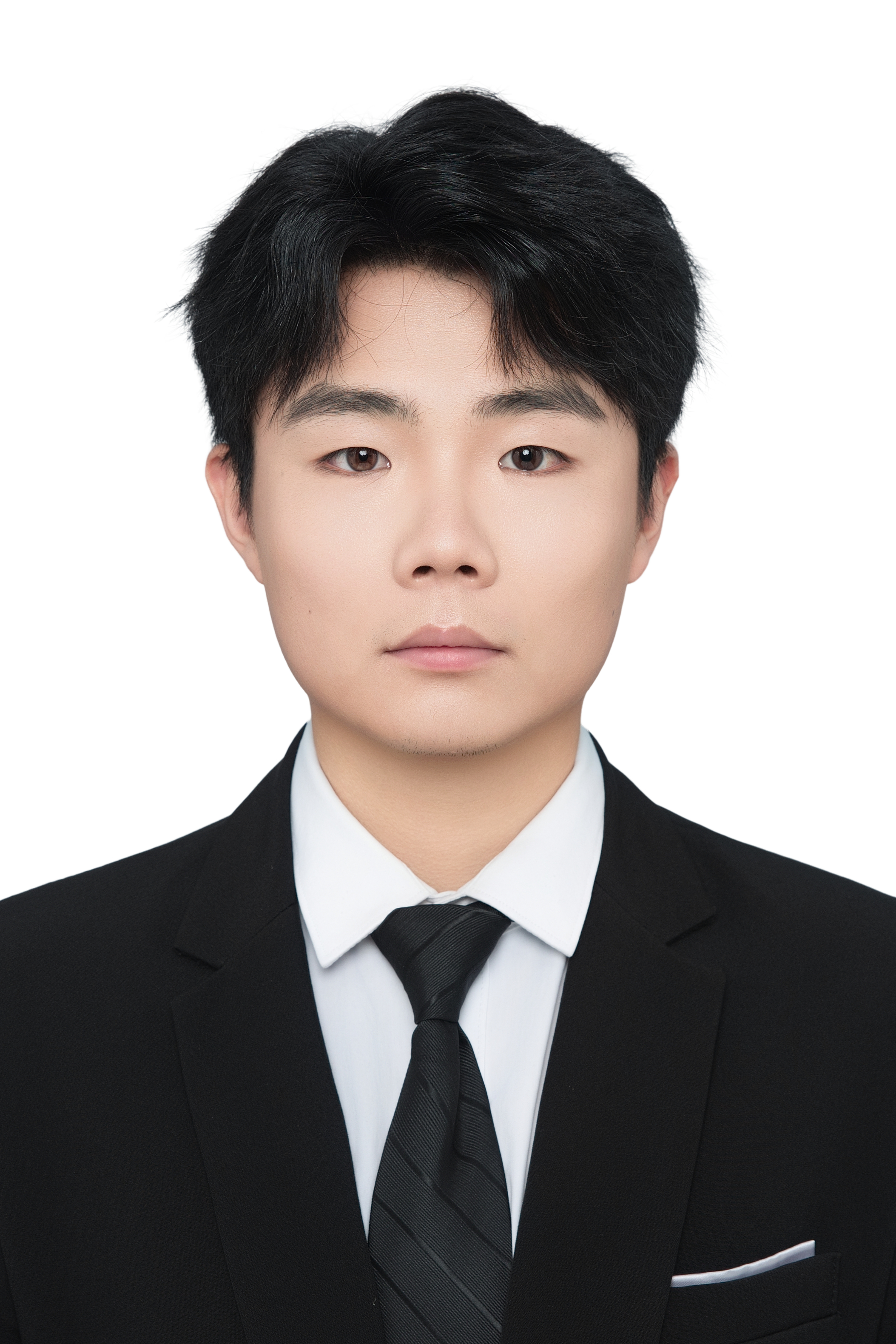}}]
{Jin Wang} received the B.S. degree in automation from Qingdao University of Science and Technology, Qingdao, PR China, in 2021. He is currently pursuing the Ph.D. degree with China University of Petroleum (East China), Qingdao, PR China. His research interests include computer vision, deep learning, and pattern recognition, especially semantic segmentation.
\end{IEEEbiography}
\vspace{-4em}
\begin{IEEEbiography}[{\includegraphics[width=1in,height=1.25in,clip,keepaspectratio]{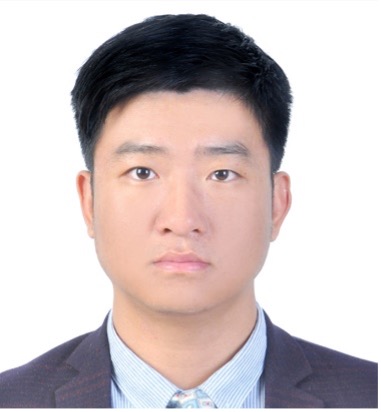}}]
{Bingfeng Zhang} received the B.S. degree in electronic information engineering from China University of Petroleum (East China), Qingdao, PR China, in 2015, the M.E. degree in systems, control and signal processing from University of Southampton, Southampton, U.K., in 2016. And the Ph.D degree from the University of Liverpool, Liverpool, U.K. His current research interest is computer vision, including semantic segmentation with limited annotation, salient object detection, and video object segmentation.
\end{IEEEbiography}
\vspace{-4em}
\begin{IEEEbiography}[{\includegraphics[width=1in,height=1.25in,clip,keepaspectratio]{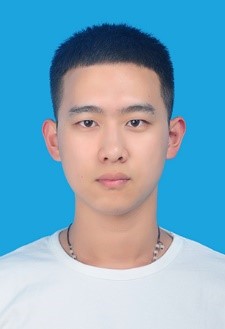}}]
{Jian Pang} received the B.S. degree in electronic information engineering from Qiqihar University, Qiqihar, PR China, in 2018, the M.E. degree in electronic and communication engineering from Kunming University of Science and Technology, Kunming, PR China, in 2021. And the Ph.D. degree from China University of Petroleum (East China), Qingdao, PR China in 2025. His main research interests include image enhancement, image re-identification, and object detection.
\end{IEEEbiography}
\vspace{-4em}
\begin{IEEEbiography}[{\includegraphics[width=1in,height=1.25in,clip,keepaspectratio]{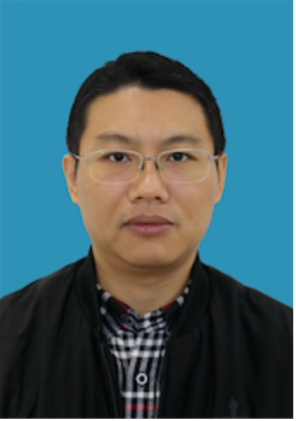}}]
{Weifeng Liu} is currently a Professor with the College of Control Science and Engineering, China University of Petroleum (East China), China. He received the double B.S. degree in automation and business administration and the Ph.D. degree in pattern recognition and intelligent systems from the University of Science and Technology of China, Hefei, China, in 2002 and 2007, respectively. His current research interests include pattern recognition and machine learning. He has authored or co-authored a dozen papers in top journals and prestigious conferences including 10 ESI Highly Cited Papers and 3 ESI Hot Papers. Dr. Weifeng Liu serves as associate editor for Neural Processing Letter, co-chair for IEEE SMC technical committee on cognitive computing, and guest editor of special issue for Signal Processing, IET Computer Vision, Neurocomputing, and Remote Sensing. He also serves dozens of journals and conferences.
\end{IEEEbiography}
\vspace{-4em}
\begin{IEEEbiography}[{\includegraphics[width=1in,height=1.25in,clip,keepaspectratio]{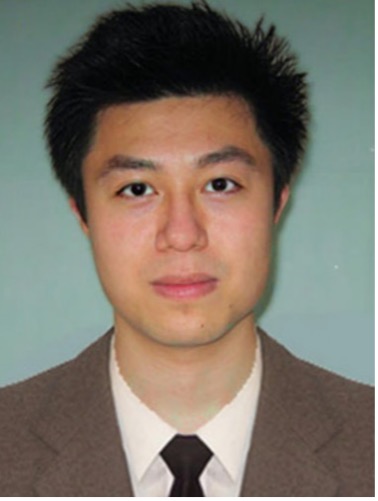}}]
{Baodi Liu} received the Ph.D. degree in Electronic Engineering from Tsinghua University. Currently, he is an assistant professor in the College of Information and Control Engineering, China University of Petroleum, China. His research interests include computer vision and machine learning.
\end{IEEEbiography}
\vspace{-4em}
\begin{IEEEbiography}[{\includegraphics[width=1in,height=1.25in,clip,keepaspectratio]{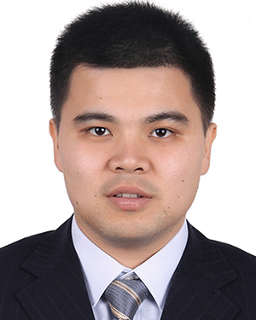}}]
{Honglong Chen} received the Ph.D. degree in computer science from The Hong Kong Polytechnic University, Hong Kong, in 2012. From 2015 to 2016, he was a Postdoctoral Researcher with the School of Computing, Informatics, and Decision Systems Engineering, Arizona State University, Tempe, AZ, USA. He is currently a Professor and a Ph.D. Supervisor with the College of Control Science and Engineering, China University of Petroleum, Qingdao, China. His current research interests include the Internet of Things and cybersecurity.
\end{IEEEbiography}

\end{document}